\documentclass[runningheads]{llncs}

\usepackage{eccv}

\usepackage{pifont}

\usepackage{eccvabbrv}
\usepackage{wrapfig}
\usepackage{graphicx}
\usepackage{booktabs}
\usepackage{enumitem}

\usepackage[accsupp]{axessibility}  
\usepackage{xcolor,colortbl}

\usepackage{hyperref}

\usepackage{orcidlink}

\expandafter\def\expandafter\normalsize\expandafter{%
    \normalsize%
    \setlength\abovedisplayskip{0pt}%
    \setlength\belowdisplayskip{8pt}%
    \setlength\abovedisplayshortskip{-8pt}%
    \setlength\belowdisplayshortskip{5pt}%
}

\begin{document}
\newcommand{\NS}[1]{\textcolor{Orange}{#1}}
\newcommand{\NSc}[1]{\textbf{\textcolor{Orange}{[RM:#1]}}}
\newcommand{\CR}[1]{\textcolor{Green}{#1}}
\newcommand{\CRc}[1]{\textbf{\textcolor{Green}{[RM:#1]}}}
\newcommand{\DP}[1]{\textcolor{cyan}{#1}}
\newcommand{\DPc}[1]{\textbf{\textcolor{cyan}{[DP:#1]}}}
\newcommand{\EV}[1]{\textcolor{purple}{#1}}
\newcommand{\EVc}[1]{\textbf{\textcolor{purple}{[OS:#1]}}}

\definecolor{myblue}{rgb}{0.19, 0.55, 0.91}
\definecolor{myred}{rgb}{0.82, 0.1, 0.26}
\newcommand{\cmark}{\textcolor{myblue}{\ding{51}}}
\newcommand{\xmark}{\textcolor{myred}{\ding{55}}}
\newcommand{\clemcolor}[1]{{\color{MidnightBlue}#1}}
\newcommand{\clem}[1]{\clemcolor{\footnotesize{}#1}}
\newcommand{\rowem}{\rowcolor{gray!20}}

\def\ours{\texttt{PAFUSE}\xspace}  
\def\mixste{\text{MixSTE}\xspace} 
\def\ddp{\text{D3DP}\xspace}

\renewcommand{\paragraph}[1]{\medskip\noindent\textbf{#1}~}

\newcommand{\inX}{\mathrm{x}}
\newcommand{\gtY}{\mathrm{y}}
\newcommand{\gtR}{\mathrm{r}}
\newcommand{\predY}{\hat{\mathrm{y}}}
\newcommand{\predR}{\hat{\mathrm{r}}}
\newcommand{\inXP}[1]{\mathrm{x}_{\text{#1}}}
\newcommand{\gtYP}[1]{\mathrm{y}_{\text{#1}}}
\newcommand{\predYP}[1]{\hat{\mathrm{y}}_{\text{#1}}}
\newcommand{\gtRP}[1]{\mathrm{r}_{\text{#1}}}
\newcommand{\predRP}[1]{\hat{\mathrm{r}}_{\text{#1}}}
\newcommand{\anyP}[1]{\bullet_{\text{#1}}}
\newcommand{\denoiser}[1]{\mathrm{D}_{\text{#1}}}
\newcommand{\pose}{\mathrm{p}}
\newcommand{\loss}{\mathcal{L}}
\newcommand{\lossWb}{\loss_{\text{WB}}}
\newcommand{\losspb}{\loss_{\text{PM}}}
\newcommand{\R}{\mathbb{R}}
\newcommand{\E}{\mathbb{E}}

\title{PAFUSE: Part-based Diffusion for 3D Whole-Body Pose Estimation} 

\titlerunning{PAFUSE}

\author{Nermin Samet\inst{1}\orcidlink{0000-0001-9247-2504} \and
C\'edric Rommel\inst{1}\orcidlink{0000-0002-9416-0288} \and
David Picard\inst{2}\orcidlink{0000-0002-6296-4222} \and
Eduardo Valle\inst{1}\orcidlink{0000-0001-5396-9868}}

\authorrunning{N.~Samet et al.}

\institute{Valeo.ai, Paris, France \and
LIGM, Ecole des Ponts, Univ Gustave Eiffel, CNRS, Marne-la-Vall\'ee, France}

\maketitle

\begin{abstract}
  We introduce a novel approach for 3D whole-body pose estimation, addressing the challenge of scale- and deformability- variance across body parts brought by the challenge of extending the 17 major joints on the human body to fine-grained keypoints on the face and hands. In addition to addressing the challenge of exploiting motion in unevenly sampled data, 
we combine stable diffusion to a hierarchical part representation which predicts the relative locations of fine-grained keypoints within each part (e.g., face) with respect to the part's local reference frame.
On the H3WB dataset, our method greatly outperforms the current state of the art, which fails to exploit the temporal information. We also show considerable improvements compared to other spatiotemporal 3D human-pose estimation approaches that fail to account for the body part specificities. Code is available at \url{https://github.com/valeoai/PAFUSE}.

 \keywords{3D Whole-body Pose Estimation \and Diffusion for Pose Estimation \and 2D to 3D Whole-body Pose Lifting}
\end{abstract}

\section{Introduction}
\label{sec:introduction}
\setlength\intextsep{8mm}
 \begin{figure}
 \centering
\includegraphics[width=0.9\linewidth]{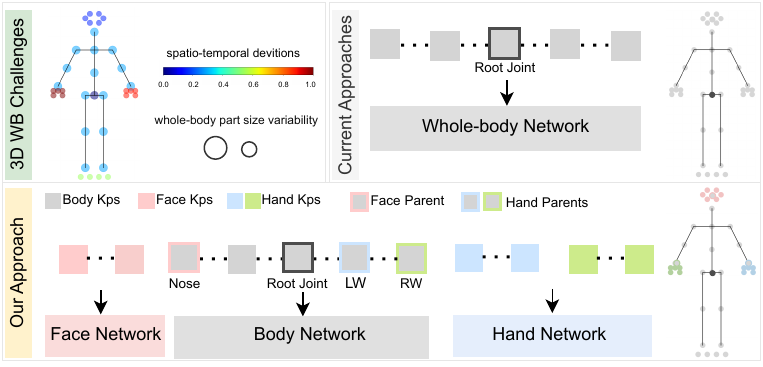}
\caption{In a whole-body skeleton, different keypoints have different scales and variations (top left) which presents a challenge for spatio-temporal prediction. Current approaches (top right) process all keypoints in a single network and, as such, have difficulties adapting to the different statistics of each body part. Our approach (bottom) groups keypoints by body parts that share similar behavior and processes them with dedicated networks, allowing better-adapted predictions.}
\label{fig:teaser}
\vspace{-7mm}
\end{figure}

As a challenging computer vision task, 3D human pose estimation aims to localize 3D human body keypoints in images and videos. 3D Human pose estimation has an important role in several vision tasks and applications such as action recognition~\cite{action1, action2, action3, action4}, human mesh recovery~\cite{meshrecovery1, meshrecovery2}, motion generation~\cite{genereation1,genereation2}, sign language~\cite{sign1, sign2, sign3, sign4}, augmented/virtual reality~\cite{augmented1, augmented2}, and robotics~\cite{robotics1, robotics2, robotics3, robotics4,robotics5}. 
In recent years, the biggest leap forward in 3D human pose estimation was including the temporal aspect and predicting entire sequences of skeletons at once~\cite{mixste,poseformer}. Indeed, adding the temporal information removes some of the depth-scale ambiguities since predicted poses also have to be compatible with the motion of the human body.

\begin{figure}
    \vspace{-0.5cm}
    \centering
    \begin{minipage}[b]{0.30\linewidth}
        \centering
        \includegraphics[width=\textwidth]{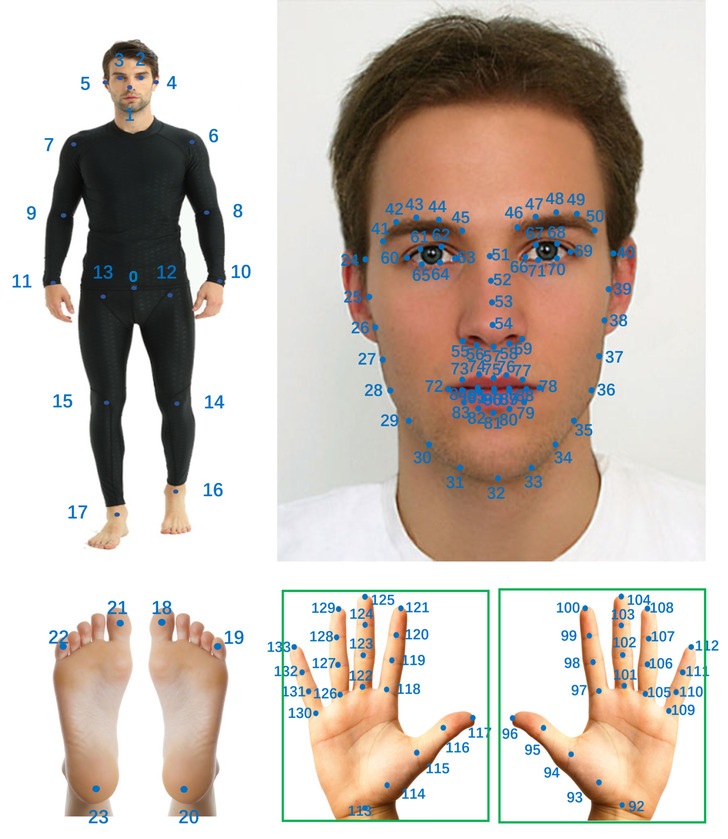}
    \end{minipage}
    \begin{minipage}[b]{0.47\linewidth}
        \centering
        \includegraphics[width=\textwidth]{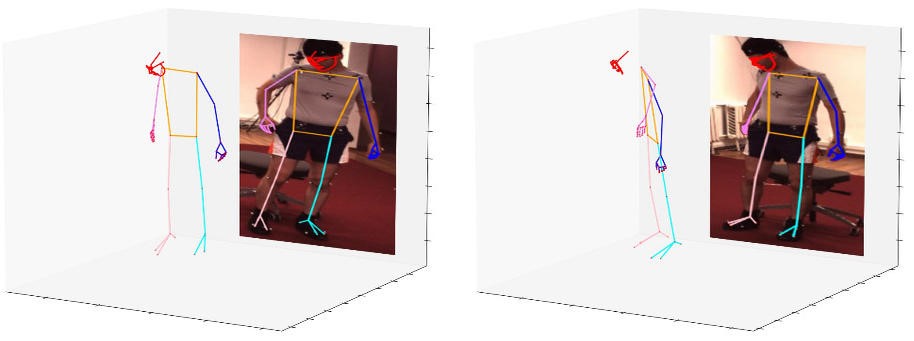}\par\vspace{5pt}
        \includegraphics[width=\textwidth]{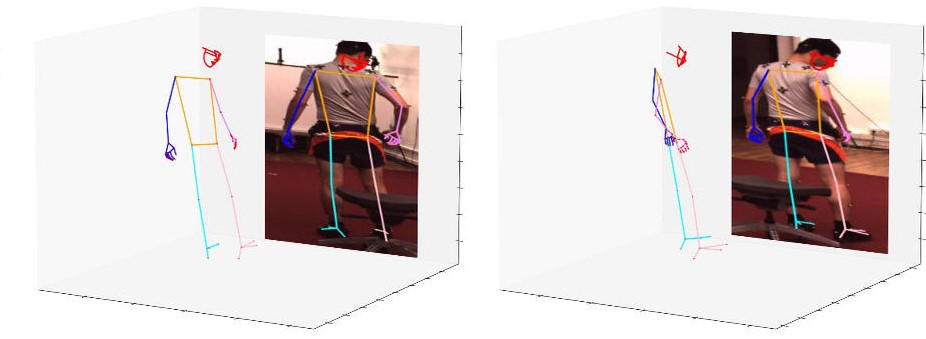}
    \end{minipage}
    \caption{(Left:) COCO-whole-body~\cite{cocowholebody} layout used in the H3WB~\cite{h3wb} dataset, with 133 keypoints. In addition to the standard 17 main-body keypoints, there are 68 face-, 42 hand- (21 keypoints for each), and 6 foot- (3 for each) keypoints. (Right:) Example 2D and 3D whole-body pose pairs from the  H3WB dataset. Images taken from~\cite{cocowholebody,h3wb}.}
    \label{fig:datasets}
    \vspace{-1cm}
\end{figure}

3D whole-body pose estimation further expands its scope by aiming to detect not only standard human body keypoints but also face, hand, and foot keypoints to enable more precise applications. 
Recently Zhu \etal~\cite{h3wb} addressed the missing dataset and benchmark issue by introducing a novel dataset for 3D whole-body pose estimation, called Human3.6M 3D WholeBody or H3WB for short. H3WB extends the Human3.6M keypoints dataset~\cite{human36m1, human36m2} by further annotating face, hand, and foot keypoints according to the layout of the COCO WholeBody dataset~\cite{cocowholebody} (see Fig.~\ref{fig:datasets}), and thus provides a unified benchmark for 3D whole-body pose estimation compatible with existing 2D whole-body methods. 
3D whole-body human pose estimation is much more challenging than regular 3D pose estimation in that the additional keypoints have different scales and diversity in poses. That is exacerbated by considering motion, since some body parts have a much greater motion range (\textit{e.g.}, hands compared to the hip, see \cref{fig:teaser} top left).

In this paper we tackle 3D whole-body human pose estimation from such temporal perspective.
Inspired by the progress in 2D whole-body pose estimation and the success of stable diffusion for 3D pose estimation, we propose a new part-based diffusion approach, \ours, that explicitly handles the aforementioned challenges. To this end, we build a 3-level 3D pose estimation network for body, hands and face. We condition each body part on their part root joint, leading to a hierarchical system. We jointly train each body part and estimate all whole-body keypoints in their specific coordinate systems (see \cref{fig:teaser} bottom). Such hierarchical bottom-up approach brings two benefits. First, the scale variance among different body parts are handled naturally as the relative distances within each part are in a similar range and each part-type is processed by a separate sub-network. Second, it brings flexibility to adjust the capacity of each sub-network according to the difficulty and deformability level of its corresponding body part. Additionally, our method is modular and can be easily added to any 3D pose estimation method to extend them to whole-body~\cite{videopose, poseformer,d3dp,mixste,mhformer,diffhpe}.

To test our ideas, we extend the H3WB~\cite{h3wb} dataset to spatio-temporal prediction. We show that our part-based approach is able to provide significant improvement to a wide variety of spatio-temporal baselines, and that our diffusion based method \ours is able to obtain state of the art results at 41.4mm MPJPE (against 88.3mm for the previous state of the art) while also generalizing to in-the-wild sequences.

To sum up, we are the first to successfully combine a part-based approach with denoising diffusion in 3D whole-body pose estimation, thus handling the variance of both scale and motion among body parts. Our contributions are the following:

\begin{itemize}[left=1em]
\item[\checkmark] We introduce a hierarchical part-based approach for 3D whole-body human pose estimation that solves the scale and motion variation issues without any additional computational cost.
\item[\checkmark] Based on this part-based approach, we propose \ours, a denoising diffusion model for 3D whole-body human pose estimation that obtains state-of-the-art performance.
\item[\checkmark] We extend the H3WB dataset to spatio-temporal prediction and perform an extensive comparison of state-of-the-art 3D human pose estimation methods on this new benchmark by adapting recent method from the literature with our hierarchical part-based approach.
\end{itemize}

\section{Related Work}
\label{sec:relatedwork}
\paragraph{3D whole-body pose estimation} 
Before the H3WB~\cite{h3wb} dataset, 3D whole-body pose estimation methods can be categorized into two groups: parametric models and distillation-based non-parametric approaches. The majority of works belong to the first category, based on parametric human body models such as Adam~\cite{DBLP:journals/corr/abs-1801-01615(totalcapture)} and SMPL-X~\cite{smplifyx}. For example, MTC~\cite{xiang2019monocular} builds upon the Adam model~\cite{DBLP:journals/corr/abs-1801-01615(totalcapture)} by optimizing its parameters after extracting 2.5D predictions. Another popular parametric model, SMPLify-X, optimizes the parameters of the SMPL-X model~\cite{smplifyx} to align with 2D keypoints. Parametric models offer the advantage of sampling nearly infinite keypoints from the mesh~\cite{afit, fieraru2020three, fieraru2021learning}. However, they have several drawbacks, including being slow and sensitive to parameter initializations. Moreover, their accuracy typically falls behind that of detection-based methods, especially on finer body parts like hands~\cite{h3wb}.

On the other hand, nonparametric methods~\cite{dope, expose, frankmocap} employ different strategies to circumvent heavy optimization procedures. Both DOPE~\cite{dope} and FrankMocap~\cite{frankmocap} initially train separate models for the body, hands, and face, which are subsequently integrated within a unified learning framework. DOPE~\cite{dope} obtains pseudo-ground annotations from these individual body models and utilizes them to supervise the distillation model. Similarly, ExPose~\cite{expose} begins by obtaining a pseudo-ground truth dataset through fitting the SMPL-X model to in-the-wild images, then proceeds to train a joint model to generate whole-body poses. One significant drawback of these methods is their reliance on different datasets for each body part. Consequently, each method produces varying whole-body layouts with differing numbers of keypoints.

Recently, Zhu \etal. utilized existing image-based 3D pose estimation methods on the H3WB dataset~\cite{h3wb} to establish a benchmark. However, as these methods do not directly tackle the challenges of 3D whole-body pose estimation, their performances are suboptimal.

\paragraph{2D Whole-Body Pose Estimation}
Following the release of the COCO WholeBody dataset~\cite{cocowholebody}, significant progress has been made in 2D whole-body pose estimation. Several methods have been proposed, with the primary focus being to overcome the challenge of scale variance in whole-body pose estimation tasks.
ZoomNet~\cite{cocowholebody} and ZoomNas~\cite{zoomnas} are among the pioneering methods in this field. They initially predict body keypoints and subsequently address scale variance by cropping the hand and face areas and transforming them to higher resolutions to further refine face and hand keypoint estimation. HPRNet~\cite{hprnet} employs a different bottom-up strategy where keypoints on each body part are separately regressed after being offset according to its part-center. TCFormer~\cite{tcformer} introduces dynamic tokenization to handle size variance in body parts by clustering tokens. Keypoint Communities~\cite{keypointcommunities} assigns different weights to body parts after constructing a skeleton graph. Seong et al.~\cite{kal} propose a keypoint-wise adaptive method to handle the scale difference between body parts. 

\paragraph{3D human pose estimation} 
We can categorize the  monocular 3D human pose estimation into two groups: deterministic and generative approaches. Early deterministic approaches, followed end-to-end pipeline to predict 3D keypoints directly from images~\cite{pavlakos2017coarse, moreno20173d, mehta2017vnect, sun2017compositional}. Later, two-stage approach become dominant where first 2D keypoints are predicted and then they are lifted into 3D using lightweight networks~\cite{sb}.
Later in order to exploit temporal information video based approaches are proposed based on convolutional neural networks~\cite{videopose} and graph convolutional networks~\cite{zhao2019semantic, liu2020comprehensive, cai2019exploiting, zou2021modulated, xu2021graph}.
More recently, attention based spatial-temporal transformer architectures gained attention such as MixSTE~\cite{mixste} and Poseformer~\cite{poseformer}.

Lifting 2D pose to 3D space is inherently ambiguous as multiple 3D poses to map onto the same 2D input. One drawback of deterministic methods lies in their inability to effectively address this ambiguity. Motivated by this challenge and limitations of deterministic methods, researchers explored multi-hypothesis approaches, including variational autoencoders~\cite{sharma2019monocular}, normalizing flows~\cite{kolotouros2021probabilistic}, and, more recently, diffusion models~\cite{diffpose, ddp, diffhpe}.
D3DP~\cite{ddp} adopts the MixSTE backbone as its denoiser and directly integrates raw 2D keypoints as conditions. In contrast to employing a straightforward averaging approach for hypothesis aggregation, it introduces a novel method based on 2D reprojections. DiffPose~\cite{diffpose} employs a diffusion based on Gaussian mixture models trained on 2D heatmaps. Additionally, apart from utilizing MixSTE as a denoiser, it employs a pre-trained MixSTE to initialize the reverse diffusion during inference. On the other hand, the denoiser architecture of DiffHPE~\cite{diffhpe} is based on CSDI~\cite{tashiro2021csdi} and TCD~\cite{saadatnejad2023generic}, where they use graph-convolutional layers to strike a good balance between computation and accuracy. During both training and inference, they utilize a pre-trained MixSTE for conditioning.

Similar to D3DP~\cite{ddp} and DiffPose~\cite{diffpose}, our proposed model \ours, also uses MixSTE as its denoiser network. We also follow~\cite{ddp, diffhpe} and incorporate raw 2D part keypoints to condition diffusion process as it is shown to be simple and very effective.

\setlength\intextsep{8mm}
 \begin{figure}
 \vspace{-0.5cm}
 \centering
\includegraphics[width=0.8\linewidth]{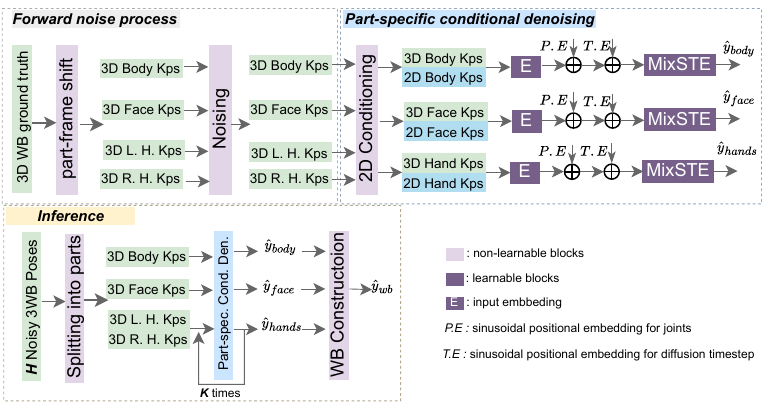}
\caption{Overall processing pipeline. During training, we split the input samples into body-part-specific tensors before performing the forward noising process (top-left). Then we train the part-based conditional denoising diffusion models (top-right). During inference (bottom), we start from random Gaussian noise and iterate the part-based conditional denoising diffusion models $K$ times to obtain the skeleton parts, which we reconstruct into a whole-body skeleton. For simplicity, we omitted the temporal aspect, although our method actually processes video sequences consisting of N frames.}
\label{fig:model}
\vspace{-1cm}
\end{figure}

\section{Method}
\label{sec:method}

We assume 2D keypoints are available and focus on lifting them to the 3D space. Thus, our task is to convert an input tensor $\inX \in \R^{2 \times J \times N}$ into a predicted output tensor $\predY \in \R^{3 \times J \times N}$, where $J$ is the number of joints (keypoints) in the reference skeleton, and $N$ is the length of the video sequence in frames. We assume that a training dataset of $(\inX,\gtY)$ pairs is available, $\gtY$ being the ground-truth 3D output. The subtensors $\anyP{body}$, $\anyP{hands}$, $\anyP{face}$ refer to the joints on the body parts indicated in the subscript. We indicate joint-wise tensor concatenation by a vertical bar, e.g., $\inX=\inXP{body}|\inXP{hands}|\inXP{face}$.

\paragraph{Overview} \cref{fig:model} shows the overall pipeline of \ours. From a sequence $\inX$ of 2D keypoints representing a whole-body pose evolving in time, we predict the corresponding 3D positions. \ours combines a part-based approach with a diffusion-based inference. The former splits the keypoints into main body, face, and hands, anchoring each part into a local frame of reference, as explained below. The latter leverages generative modelling, separately on each part, to propose multiple pose hypotheses. The combined approach allows addressing both the depth-ambiguity inherent to monocular human-pose estimation, and the specific needs of body parts with very different scales, motions, and deformabilities.

\paragraph{Whole-Body-Frame vs. Part-Frame Shift.} Usually, $\gtY$ is represented in a way that it sets the body root joint (keypoint 0, at the center of the hips) at the origin of the coordinate system. We propose an alternative representation, with different body parts centered at local coordinate frames: keypoint 0 for the main body, the nose (keypoint 1) for the face, and the corresponding wrists (keypoints 10 and 11) for each hand.

\paragraph{Generative-based prediction.} Our learning engine is a denoising diffusion probabilistic models (DDPM)~\cite{ddpm}, typically employed for generative tasks, but here leveraged to model the distribution $\Pr(\gtY|\inX)$. DDPMs are trained to reverse the incremental addition of Gaussian noise to data, until they are able to ``restore'' convincing data samples from pure Gaussian noise.

\textit{Forward noise process.}
Following the usual DDPM procedure, the diffusion process samples a ``timestep'' $t \sim \text{U}(0, T)$ and then corrupts ground-truth 3D tensor by adding a Gaussian noise whose intensity increases with $t$, \ie, $\gtY^t = \gtY + \epsilon^t$, $\epsilon^t \sim \mathcal{N}(0, f(t)I)$, where $f$ is a monotonically increasing function of $t$, which is designed according to the desired noising schedule. It should be clear that the idea of ``time'' expressed by $t$ refers to the noising process and has nothing to do with the pose motion: the entire pose-in-motion tensor (with its $N$ frames) is noised and denoised all at once. In this work, we did not optimize the noise scheduler and use a popular cosine noise scheduler~\cite{nichol2021improved}.

\textit{Part-specific conditional denoising.}
In \ours, we learn independent denoising models for the main body, the face, and the hands. At the denoising step, we split the noised ground-truth, such that $\gtY^t = \gtYP{body}^t | \gtYP{face}^t | \gtYP{L-hand}^t | \gtYP{R-hand}^t$. The denoising is conditioned by the timestep $t$, which we positionally-encode with a family of sinusoidal functions before feeding it to the decoder (similar to how token positions are encoded in transformers). Crucially, the denoising of each part is also conditioned on the corresponding input 2D tensor (\eg, $\inXP{face}$ for $\gtYP{face}^t$), by simply concatenating the tensors across the space dimension (\ie, conditioned inputs $\in \R^{5 \times J_{\text{part}} \times N}$). Note that input 2D tensors are used as it is,  \textit{i.e.}, no offsetting for each part.
Although DDPM denoising is often learned stepwise, by asking the denoiser to reconstruct $\gtY^{t-1}$ from $\gtY^t$, here we follow D3DP~\cite{d3dp} and learn to reconstruct the denoised input directly:

\begin{equation}
\predYP{part} = \denoiser{part}(t, \inXP{part}, \gtYP{part}^t)
\label{eq:denoising_prediction}
\end{equation}
\noindent where $\denoiser{part}$ is the part-specific denoising network.

\textit{Conditioned inference.}

Test time inference predicts $\predY$ by concatening the 2D input tensor $\inX$ to a pure-noise tensor sampled from a Gaussian distribution $\predY^T \sim \mathcal{N}(0, I)$. Although the training objective is one-step denoising (\cref{eq:denoising_prediction}), inference still requires incremental denoising ($\predY^T \rightarrow \predY^{T-1} \rightarrow \cdots \rightarrow \predY$) to produce high-quality samples from the expected conditional distribution. Following D3DP, we employ DDIM~\cite{ddim} to bridge the gap between the one-step results produced by the learned denoiser and the high-quality incremental denoising needed as final results.

\paragraph{Losses.} We experimented with two losses. The \textit{part loss} considers the keypoints in their local (part-based) frames of reference. Because the denoisers are trained in those local frames, that corresponds to computing the loss between the reassembled keypoint tensors, without further processing:

\begin{equation}
\loss_\text{part} = \ell\left(\,
    \predYP{body}|\predYP{hands}|\predYP{face}, \:
    \gtYP{body}|\gtYP{hands}|\gtYP{face} \,\right)
\label{eq:loss_part}
\end{equation}
\noindent where $\ell$ is either the MPJPE metric or the mean-squared error (MSE).

The whole-body or the \textit{WB loss} uses the whole-body frame of reference, and thus, must recover the global coordinates from the estimated root-joint positions of each local coordinates. Let $\gtR$ be the tensor containing the position (relative to keypoint 0) of the local coordinates of each joint, such that  $\gtYP{wb-frame}=\gtYP{part-frame}+\gtR$ and $\predYP{wb-frame}=\predYP{part-frame}+\predR$. Then, the WB loss is: 

\begin{equation}
\begin{aligned} 
\loss_\text{WB} = \ell(\,
    \predYP{body}+\predRP{body}|\predYP{hands}+\predRP{hands}|\predYP{face}+\predRP{face}, \\
    \gtYP{body}+\gtRP{body}|\gtYP{hands}+\gtRP{hands}|\gtYP{face}+\gtRP{face} \,)
\end{aligned}
\label{eq:loss_wb}
\end{equation}
\noindent where $\ell$ is, again,  either the MPJPE or the MSE.

Our experiments show that the part loss works better than the WB loss, and that the variant with $\ell=\text{MPJPE}$ works better than $\ell=\text{MSE}$ (\cref{sec:ablations}).

\paragraph{Novelty.} The \textit{part-based design}, associating the use of local frames of reference for each body part and the use of separate part-based models are the main novelty of \ours, in addition to the application of those techniques to a \textit{diffusion-based inference}. Our results show that our part-based design systematically improves techniques and results in state-of-the-art results when associated to diffusion.

\section{Results}
\label{sec:experiments}

\subsection{Experimental setup}

\paragraph{Dataset.} 
Despite its growing importance, 3D whole-body pose estimation has only recently been addressed by the computer-vision community, and there is only one annotated video whole-body dataset publicly available, the Human 3.6M WholeBody (H3WB) dataset~\cite{h3wb}, which is a reannotation of the Human 3.6M (H36M) dataset~\cite{human36m1,human36m2}. It contains $10^5$ ground-truth triplets of frames, 2D coordinates and 3D coordinates in camera space obtained from subjects S1, S5, S6, S7, S8; with 80\% of triplets for training (S1, S5, S6, S7) and 20\% for testing (S8). Remark that, while the frames of H36M are evenly sampled, the annotated frames of H3WB are not, posing additional challenges.
Our experiments follow the H3WB benchmark protocol and use the ground-truth 2D keypoints provided in the H3WB dataset (Fig. ~\ref{fig:datasets}). For in-the-wild videos, we extract 2D whole-body keypoints with OpenPifPaf~\cite{pifpaf}.

Since H3WB is unique in literature, we sought ways to mitigate the use of single dataset in our experiments and provided results on challenging in-the-wild videos featuring intense occlusions and motion blur.

\label{sec:experimental_setup}
~\par\vspace{-\baselineskip}
\setlength\intextsep{8mm}
\begin{wrapfigure}[18]{r}{0.5\textwidth}
\captionsetup[subfigure]{labelformat=empty}
\centering
\includegraphics[width=1\linewidth]{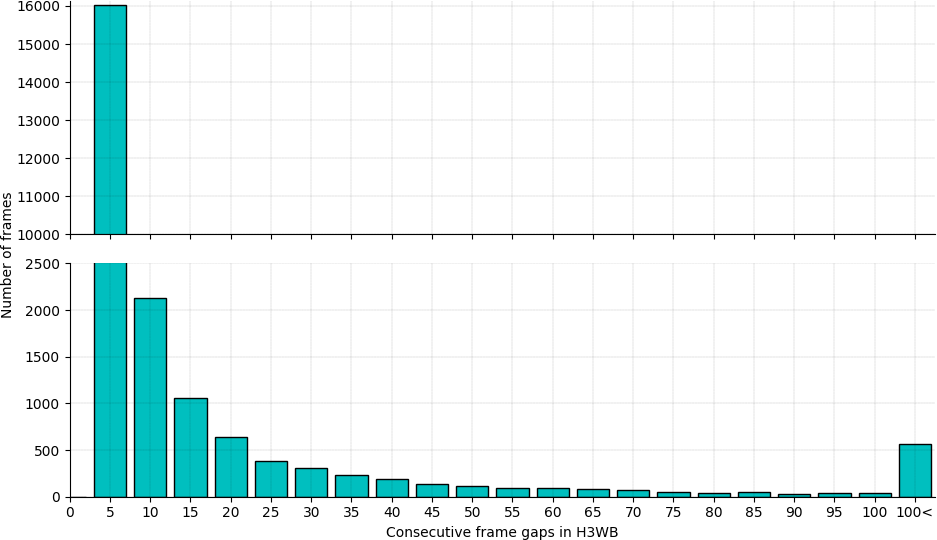}
\caption{Distribution of gaps between annotated frames of H3WB dataset, showing a long tail with many gaps of 100 or more frames. Contrast to Human3.6M, which is evenly annotated at every 5 frames. Please remark the discontinuity at the y-axis, to make room for the mode at 5 frames.}
\label{fig:uneven}

\end{wrapfigure}

\paragraph{Metrics.}
Our main evaluation metric is the mean-per-joint-position error (MPJPE), which dominates the literature of 3D human pose estimation. We compute it across all whole-body keypoints, following H3WB's official benchmark, after translating the root joint to its accurate position (``Protocol \#1''). That metric appears as \textbf{WB} in our tables. We further compute part-specific MPJPE scores for the \textbf{Body}, \textbf{Hands}, and \textbf{Face}, by considering only the keypoints of those parts, after aligning them to their respectively local root joints (keypoints 0, 1, 10 and 11, respectively for body, face, left hand and right hand). Finally, the average of those part-specific MPJPE appears as \textbf{PB} in our tables.

Following the usual practice for generative-based models, including our baseline D3DP~\cite{d3dp}, we mainly evaluate our method by selecting the hypothesis that most closely aligns with the ground truth (\textbf{P-Best}). We also employ an averaged hypothesis (\textbf{P-Agg}) as an auxiliary additional metric.

\paragraph{Training and inference details}
We employ the AdamW optimizer, with momentum parameters $\beta_1=0.9$ and $\beta_2=0.999$, and a weight decay of 0.1. Consistent with MiXSTE~\cite{mixste}, we train for 400 epochs, starting with an initial learning rate of $6\times10^{-5}$.
Due to the uneven sampling of H3WB dataset, we limited the experiments to frame windows of $N=27$ and $81$, shorter than the typical maximum of $N=243$ found in the 3D human pose estimation literature. For $N=27$, we set the batch size to $36$, and for $N=81$, to $12$. For training, we set the number of hypotheses $H=1$, and the sampling iterations $K=1$. For inference, we evaluate variations on those settings ($H=1$, $20$, and $300$, and $K=1$ and $5$). Unless otherwise specified, we trained \ours on the MPJPE-based part loss (\cref{eq:loss_part}), sequence length $N=27$, a single network for both hands, and MixSTE~\cite{mixste} denoiser backbones with 384, 256 and 224 channels, respectively, for the body, hands, and face.

Following D3DP~\cite{d3dp}, we implement \ours in PyTorch.  
We train our models on a single NVidia A100 40 GB GPU for 400 epochs, which takes about 18 hours.

\begin{table}[!htb]
    \centering
    \setlength{\tabcolsep}{5pt}
    \begin{tabular}{l r r r r r}

    \toprule
    \multicolumn{1}{l}{Method} & WB & PB &
    \multicolumn{1}{c}{Body} &
    \multicolumn{1}{c}{Face} &
    \multicolumn{1}{c}{Hands} \\

    \midrule
    \textit{Single-frame} & & & & & \\
    SMPL-X\cite{smplifyx} & 188.9 & 55.3 & 166.0 & 23.7 & 44.4 \\
    CanonPose\cite{canonpose} & 186.7 & 58.1 & 193.7 & 24.6 & 48.9   \\
    SimpleBaseline \cite{sb} & 125.4 & 45.5 & 125.7 & 24.6 & 42.5 \\
    CanonPose\cite{canonpose} \textit{with 3D supervision} & 117.7 & 39.5 & 117.5 & 17.9 & 38.3   \\
    Large SimpleBaseline\cite{sb} & 112.3 & 34.9 & 112.6 & 14.6 & 31.7 \\
    Jointformer\cite{jointformer} & 88.3 & 36.0 & 84.9 & 17.8 & 43.7  \\

    \midrule
    \textit{Spatio-temporal} & & & & & \\

    Videopose ($N=27$) & 70.1 & 27.8 & 62.2 & 11.5 & 34.7 \\
    \qquad\clem{+our part-based design} & \clem{68.4} & \clem{22.4} & \clem{57.2} & \clem{7.9} & \clem{26.1} \\
    Videopose ($N=81$) & 71.7 & 28.9 & 64.9 & 11.8 & 36.1 \\
    \qquad\clem{+our part-based design} & \clem{70.5} & \clem{22.7} & \clem{57.2} & \clem{8.1} & \clem{26.5} \\ 
    Poseformer ($N=27$) & 59.8 & 21.6 & 53.4 & 7.5 & 26.4 \\
    \qquad\clem{+our part-based design} & \clem{58.0} & \clem{18.2} & \clem{49.7} & \clem{5.7} & \clem{20.6} \\
    Poseformer ($N=81$) & 68.7 & 28.7 & 62.4 & 12.4 & 36.6 \\
    \qquad\clem{+our part-based design} & \clem{59.0} & \clem{18.9} & \clem{47.9} & \clem{6.3} & \clem{22.6} \\
    MixSTE ($N=27$) & 55.3 & 21.1 & 49.5 & 8.3 & 25.5 \\
    \qquad\clem{+our part-based design} & \clem{52.8} & \clem{20.2} & \clem{45.2} & \clem{8.1} & \clem{24.1} \\
    MixSTE ($N=81$) & 54.5 & 20.7 & 48.1 & 8.0 & 25.7 \\
    \qquad\clem{+our part-based design} & \clem{50.6} & \clem{19.8} & \clem{46.1} & \clem{7.7} & \clem{24.3} \\

    \midrule
    \textit{Spatio-temporal + diffusion} & & & & & \\

    D3DP($N=27$, $H=1$, $K=1$) & 50.9  &  20.1  &  46.3   &  8.2  & 24.3 \\
    \rowem \ours ($N=27$, $H=1$, $K=1$) & 45.6 & 16.7  & 37.8 & 6.1  & 21.9  \\

    D3DP($N=27$, $H=20$, $K=10$, P-Agg) & 49.9  & 19.6  &  45.5 & 7.8  & 23.8  \\
        D3DP($N=27$, $H=300$, $K=5$, P-Agg) & 50.1 & 19.4 & 45.2 & 7.6 & 23.6 \\

    \rowem \ours ($N=27$, $H=20$, $K=10$, P-Agg) & 45.5 & 16.6 & 37.4 & 5.8 & 22.3  \\    
    \rowem \ours ($N=27$, $H=300$, $K=5$, P-Agg) & 45.6 & 16.6 & 37.5 & 5.9 & 21.9  \\

    D3DP($N=27$, $H=20$, $K=10$, P-Best) & 46.9 & 19.4  &  44.6 & 7.9  & 23.6  \\
   
    D3DP($N=27$, $H=300$, $K=5$, P-Best) & 45.3 & 18.6 & 43.0 & 7.5 & 22.5 \\

    \rowem \ours ($N=27$, $H=20$, $K=10$, P-Best) & 43.0 & 16.4 & 37.1 & \textbf{5.8} & 21.7  \\
    \rowem \ours ($N=27$, $H=300$, $K=5$, P-Best) & \textbf{41.4} & \textbf{15.9} & \textbf{36.3} & 5.9 & \textbf{20.5}  \\

 \bottomrule
\end{tabular}

\caption{State-of-the-art comparison for 2D$\rightarrow$3D lifting on the H3WB dataset (MPJPE in mm). Despite H3WB's irregular frame sampling, spatio-temporal methods consistently improve results, as does our \clemcolor{part-based design}. \ours, combining multiple generative hypothesis and our part-based design has the best results. \textbf{The part-based models have the same total number of parameters as their vanilla counterparts.}}
\label{tab:sota}
\vspace{-1cm}
\end{table}

\subsection{Comparison with the state-of-the-art} 
\label{sec:sota_comparison}

\paragraph{Baselines.}
Single-frame methods currently dominate the state of the art on H3WB~\cite{canonpose, sb, jointformer}.
We added to those baselines well-established spatio-temporal 3D human pose estimation methods, selecting those that allow reproducible results~\cite{videopose, poseformer, mixste}. The most important baseline is D3DP~\cite{d3dp} which is both spatio-temporal and diffusion-based, but not part-based. We adapted all baselines to accept the 133 keypoints of whole-body poses, and trained them according to their official instructions. The results of the state-of-the-art comparison appear in \cref{tab:sota}.

\paragraph{Spatio-temporal consistently outcompete single-frame.} Despite the irregular frame-sampling of H3WB (see ~\cref{fig:uneven}), which poses challenges for spatio-temporal methods, they still systematically outcompete single-frame ones. However, due to that irregular sampling, all methods favor short windows ($N=27$), instead of the long windows ($N=81$) traditionally preferred in literature~\cite{videopose, d3dp, diffhpe, diffpose}. Indeed, with such long windows, some of the sequences will have very large gaps between the frames leading which makes them virtually impossible to learn because of their rarity in the training set and the lack of correlation between the frames that are to be predicted.

\paragraph{Our part-based design consistently improves techniques.} Our part-based design consistently improves all spatio-temporal techniques, in all metrics, sometimes by several mm, as shown in the \clemcolor{highlighted lines} of \cref{tab:sota}. Remark that, for a fair comparison, we used the same total number of parameters in the part-based models as for the original models, which means that there is no additional memory cost for our part-based design. The most dramatic improvements appear, as expected, on hands and face, but the body keypoints also benefit from the part separation.

\paragraph{\textit{Precise} prediction of keypoints outcompete body \textit{mesh} generation.} 
The results demonstrate that accurately predicting keypoints for the body, face, and hands is more effective than generating body meshes. SMPL-X recorded a MPJPE of 188.9, whereas all keypoint prediction methods outperformed SMPL-X. Our method, in particular, achieved a significantly lower MPJPE of 41.4.

\paragraph{PAFUSE obtains the best results.} D3DP showcases that the multiple hypotheses of diffusion allow locating better poses than single-hypothesis techniques. However, the combination of diffusion and part-based of PAFUSE is the one that presents the best results, on all metrics.

The closest method to ours, DOPE~\cite{dope}, based on distillation across different body parts, was omitted from our benchmark due to failing to consistently predict complete skeletons on H3WB, often missing occluded parts, thus precluding meaningful scores.

\begin{figure}
 \captionsetup[subfigure]{labelformat=empty}
\centering
\setlength\tabcolsep{1.5pt} 
\resizebox{0.95\textwidth}{!}{
\begin{tabular}{ccc}
\tiny{Input} & \tiny{D3DP} & \tiny{\ours}  \\

\includegraphics[width=0.12\textwidth,keepaspectratio,] {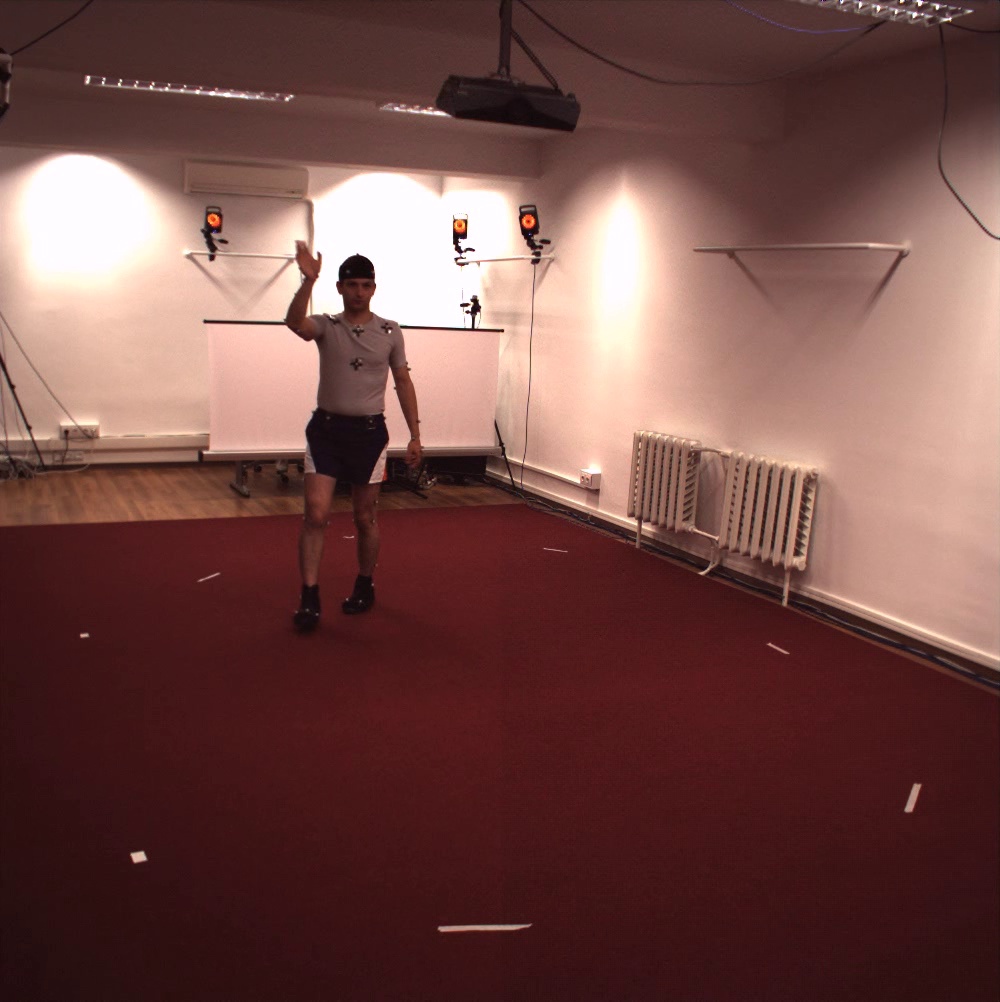} & 
\includegraphics[width=0.18\textwidth,keepaspectratio,] {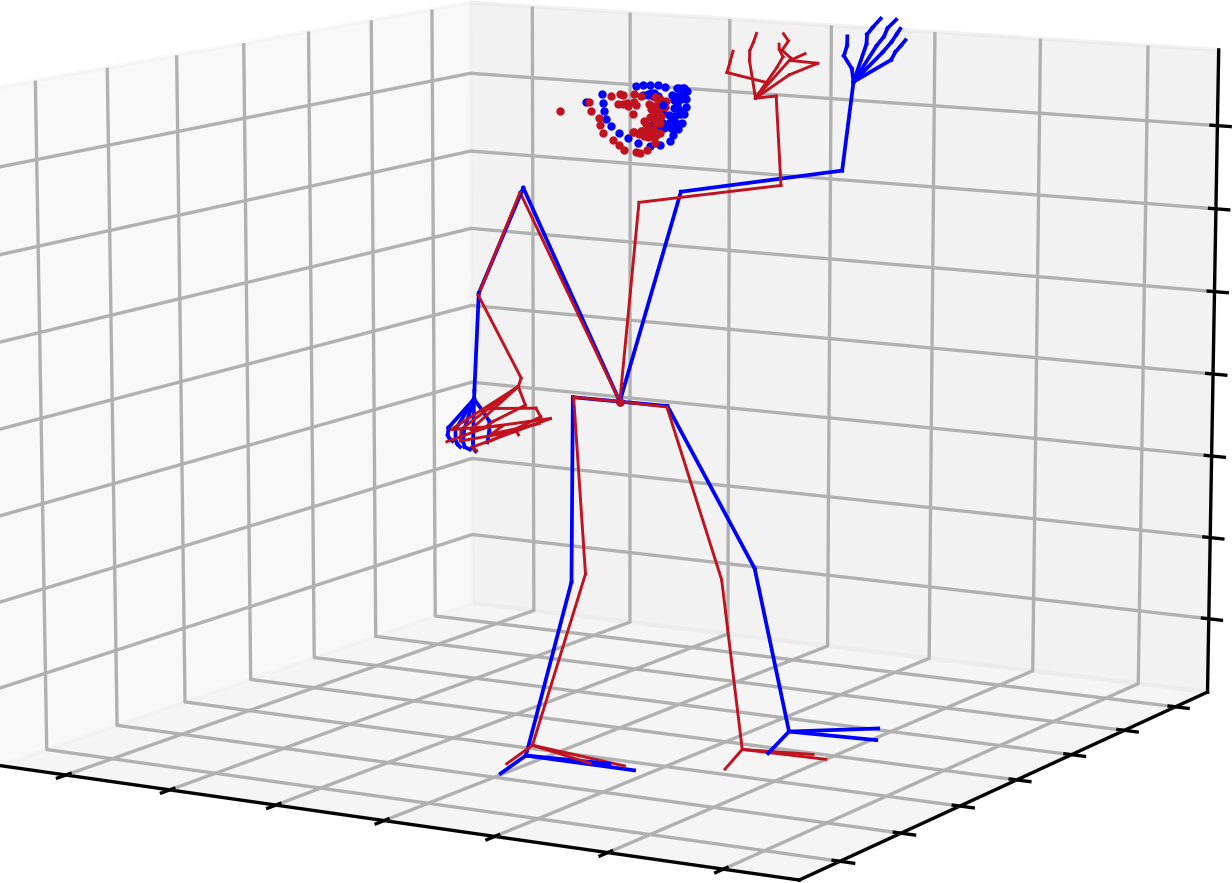} &
\includegraphics[width=0.18\textwidth,keepaspectratio,] {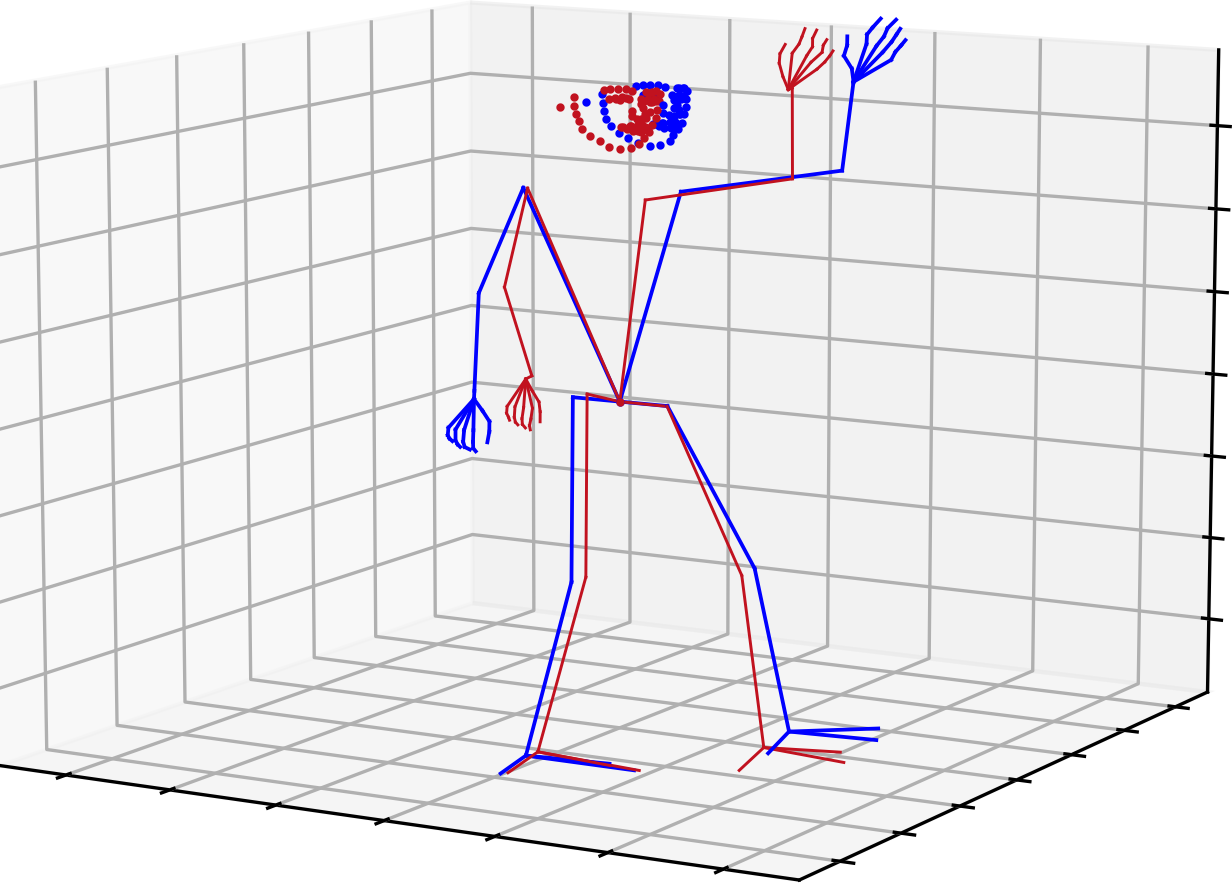} \\

\includegraphics[width=0.12\textwidth,,keepaspectratio,] {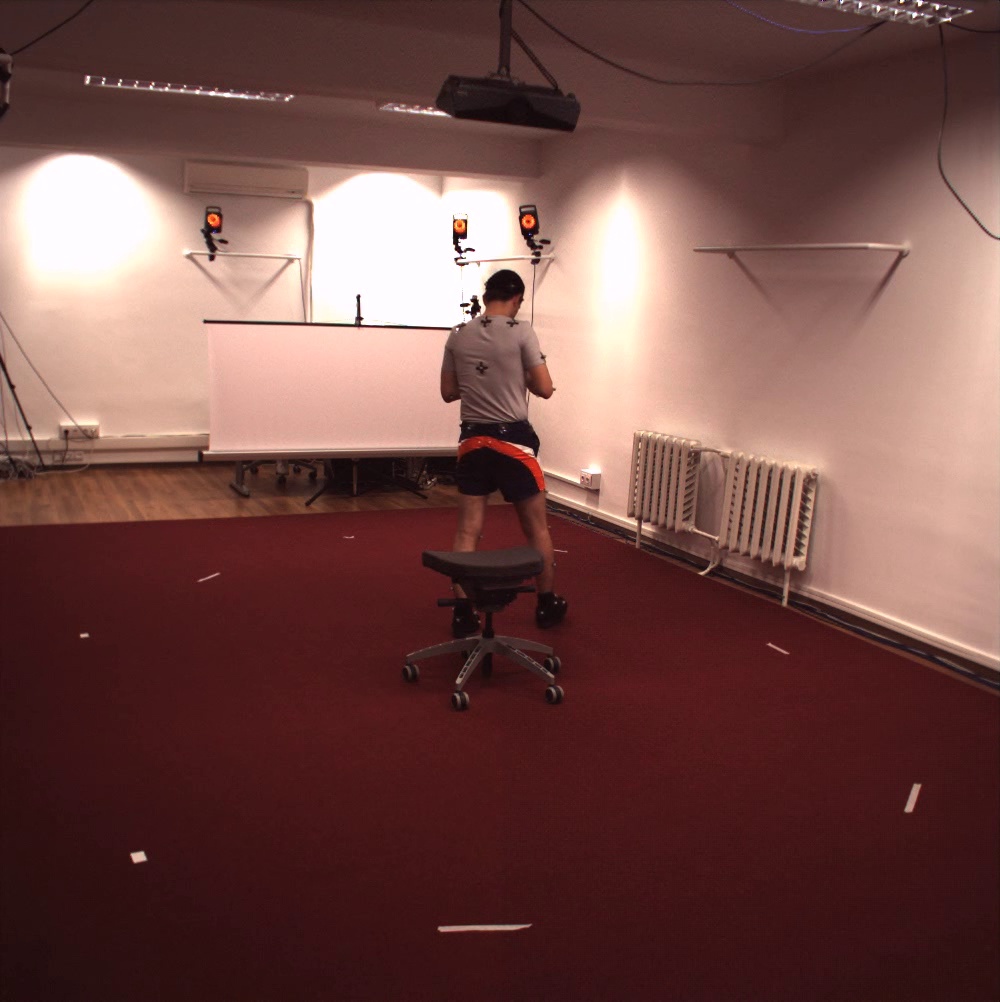} & 
\includegraphics[width=0.18\textwidth,keepaspectratio,] {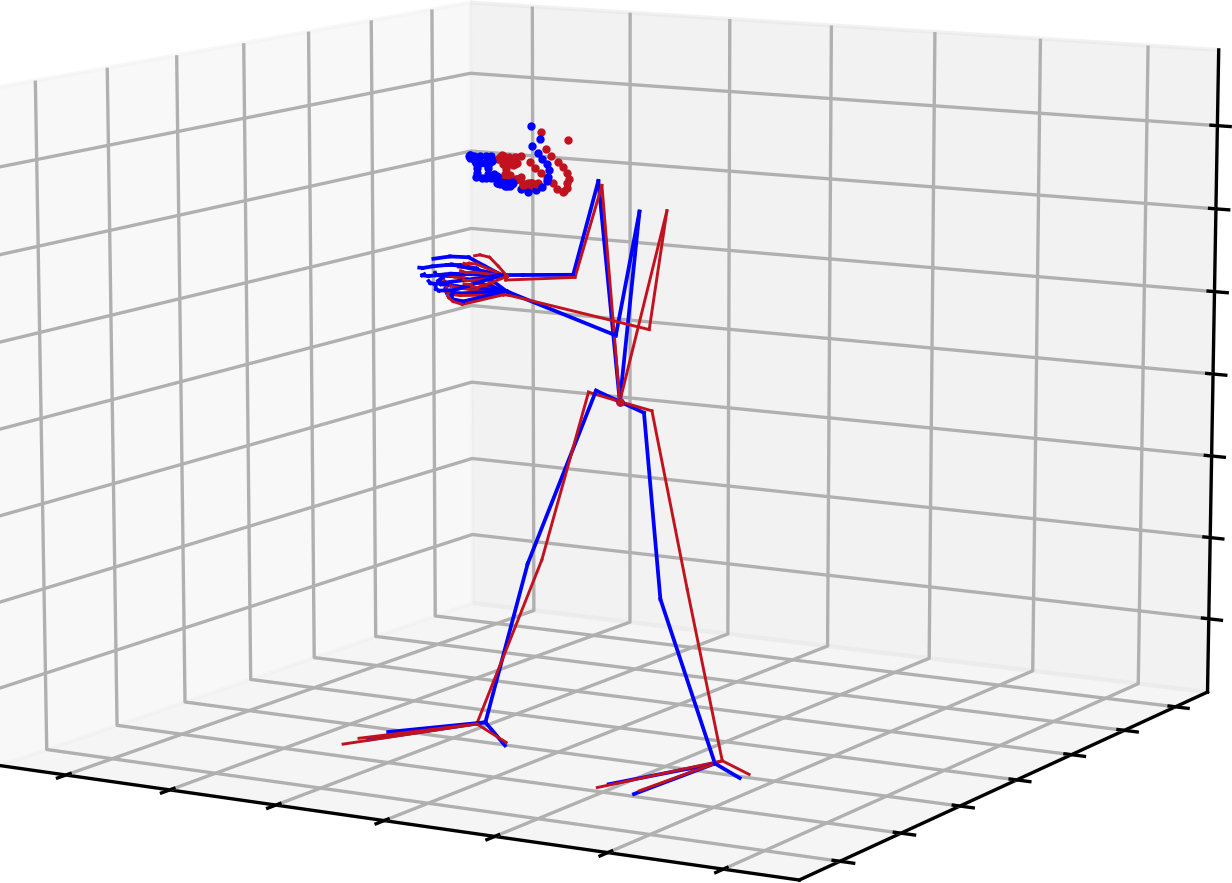} & 
\includegraphics[width=0.18\textwidth,keepaspectratio,] {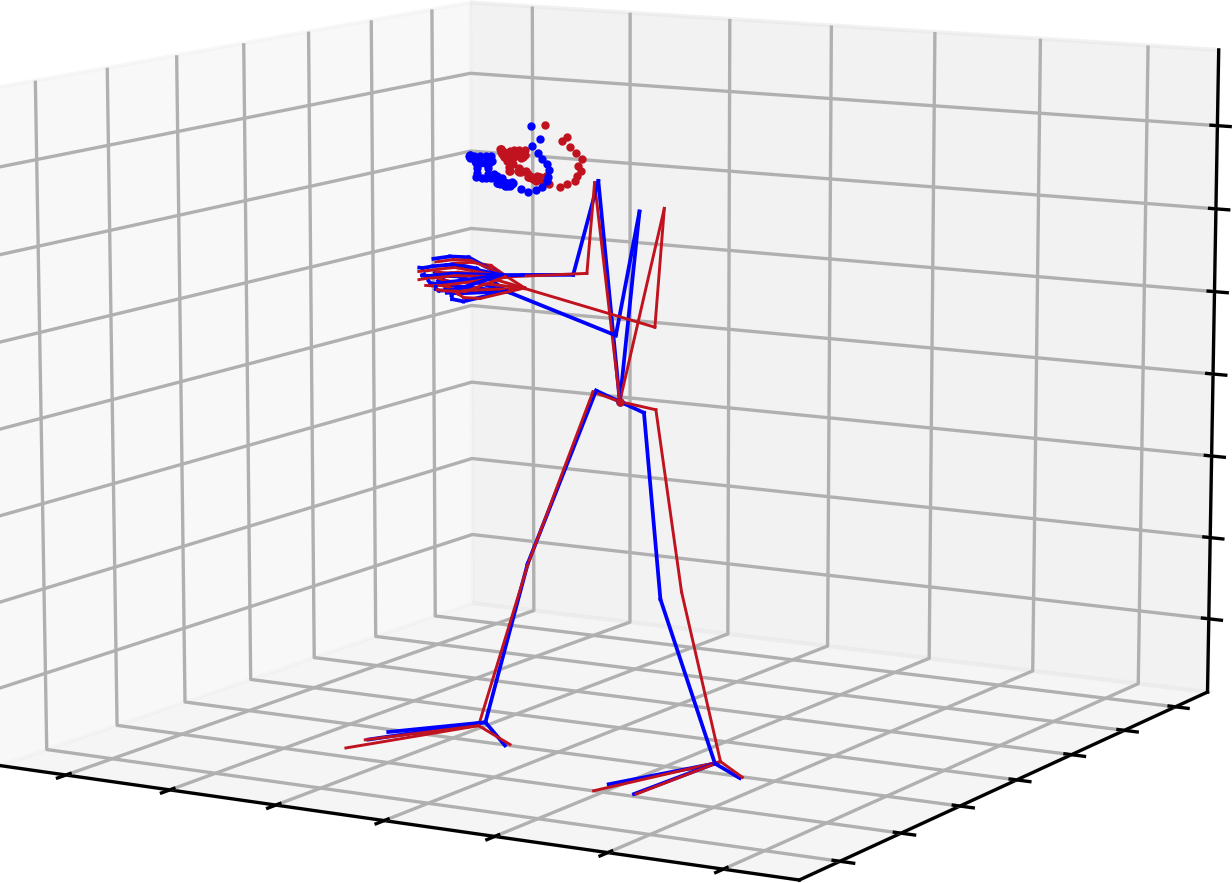} \\

\includegraphics[width=0.12\textwidth, keepaspectratio,] {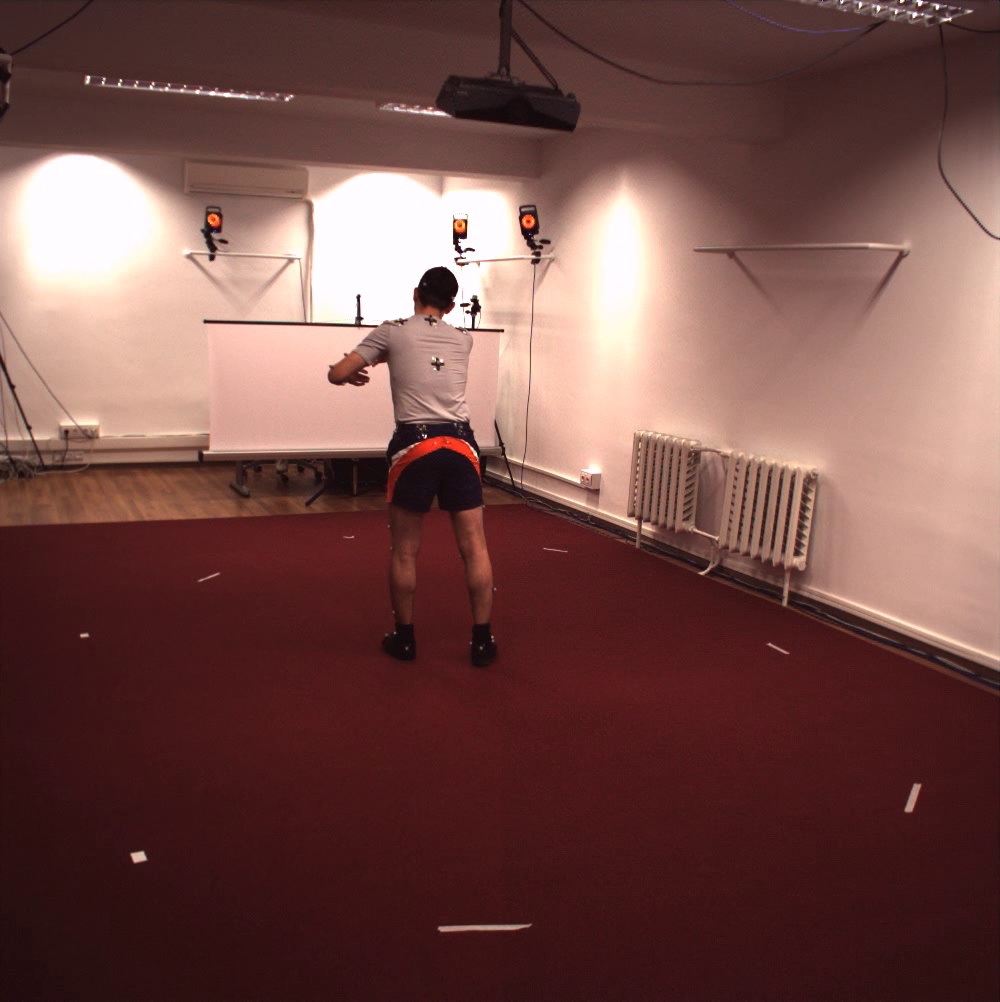} &
\includegraphics[width=0.18\textwidth,keepaspectratio,] {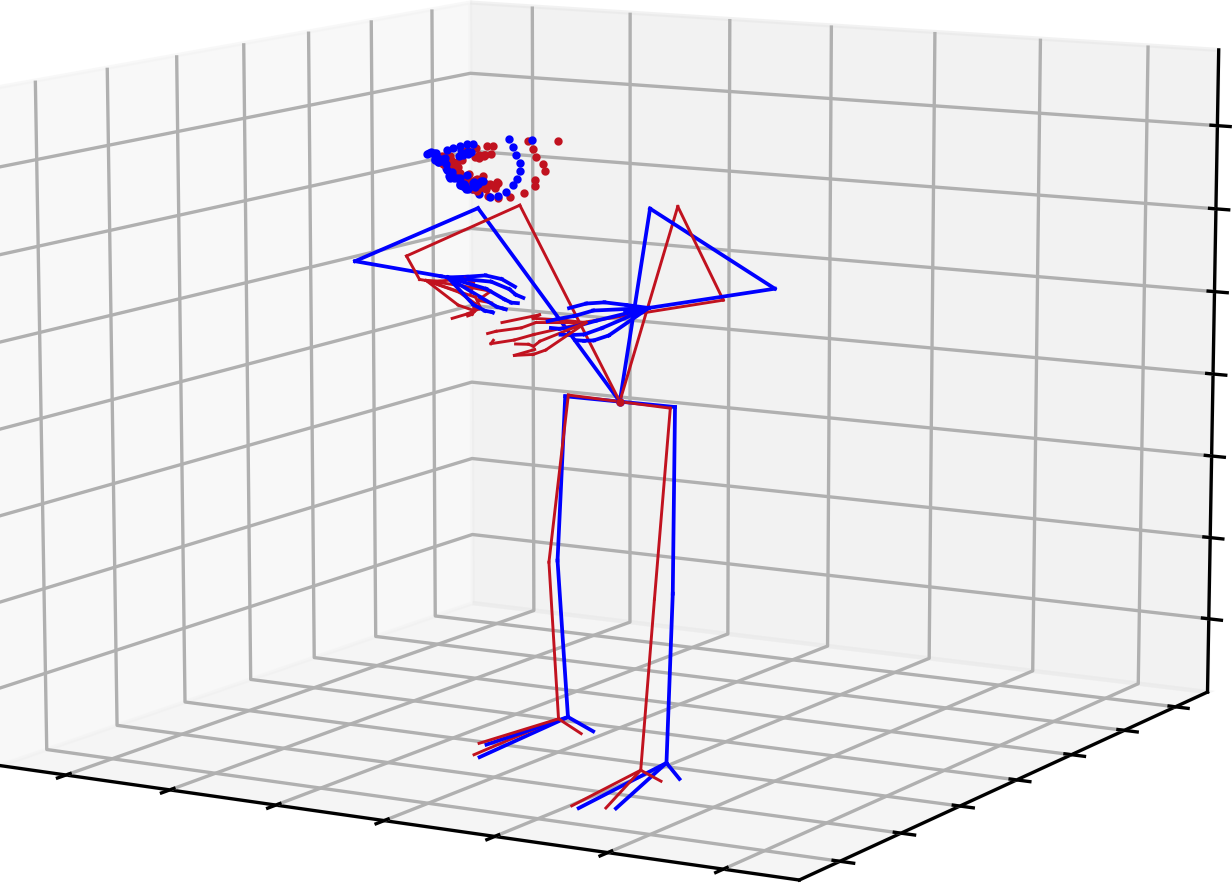} & 
\includegraphics[width=0.18\textwidth,keepaspectratio,] {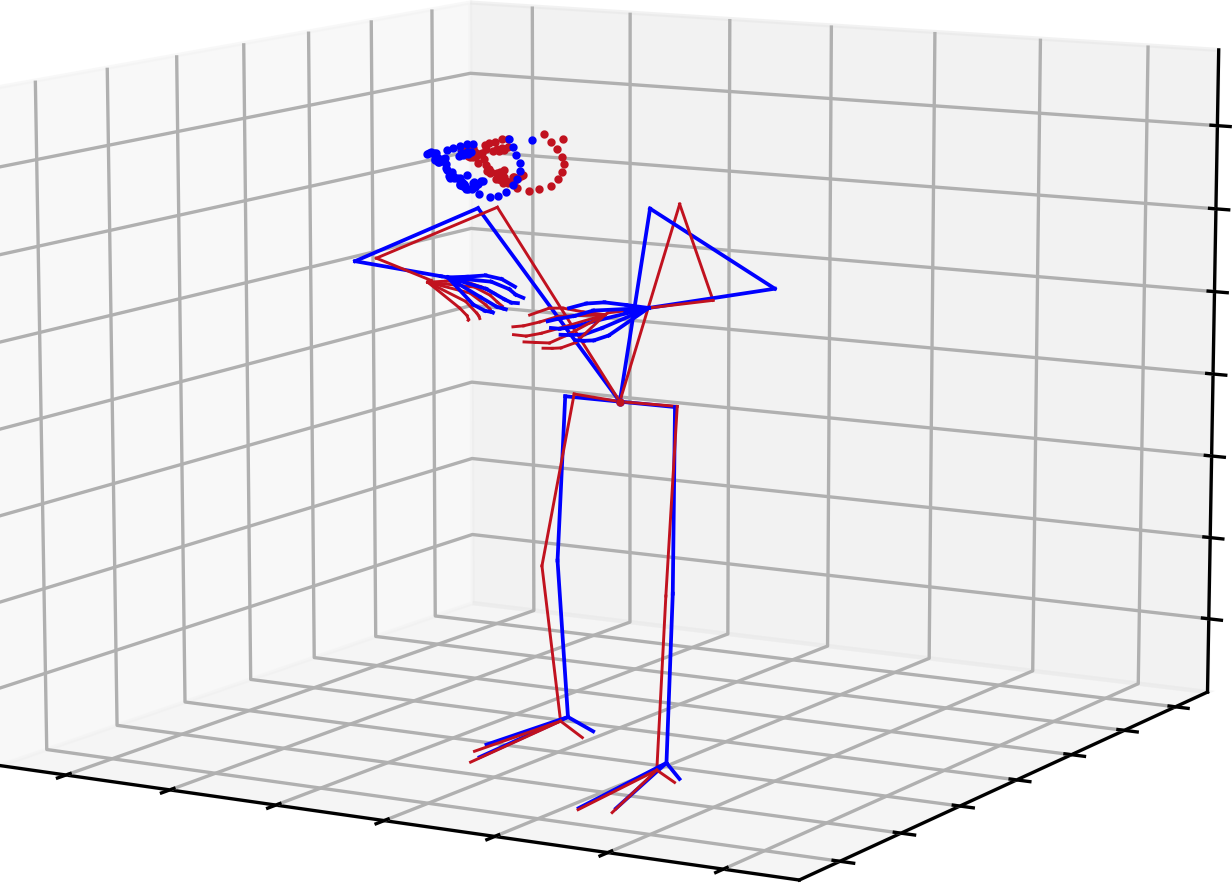} \\

\includegraphics[width=0.12\textwidth,keepaspectratio,] {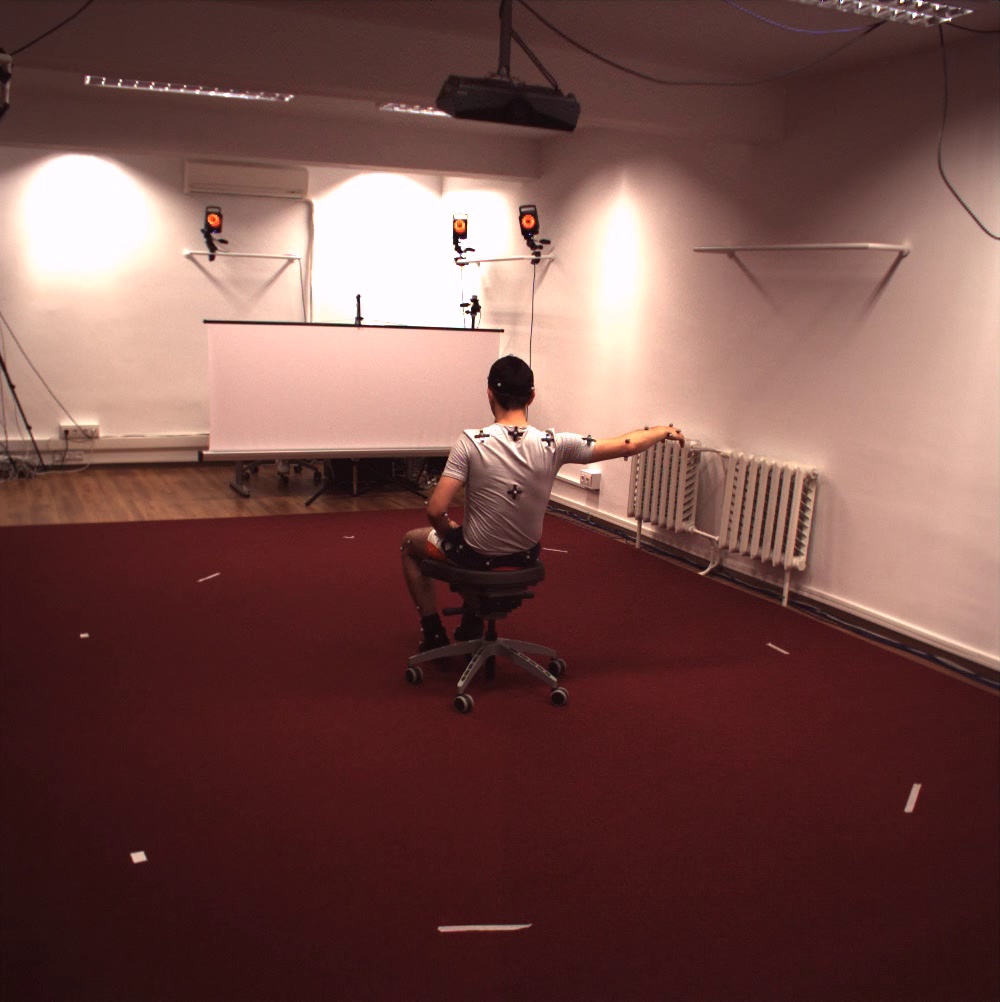} & 
\includegraphics[width=0.18\textwidth,keepaspectratio,] {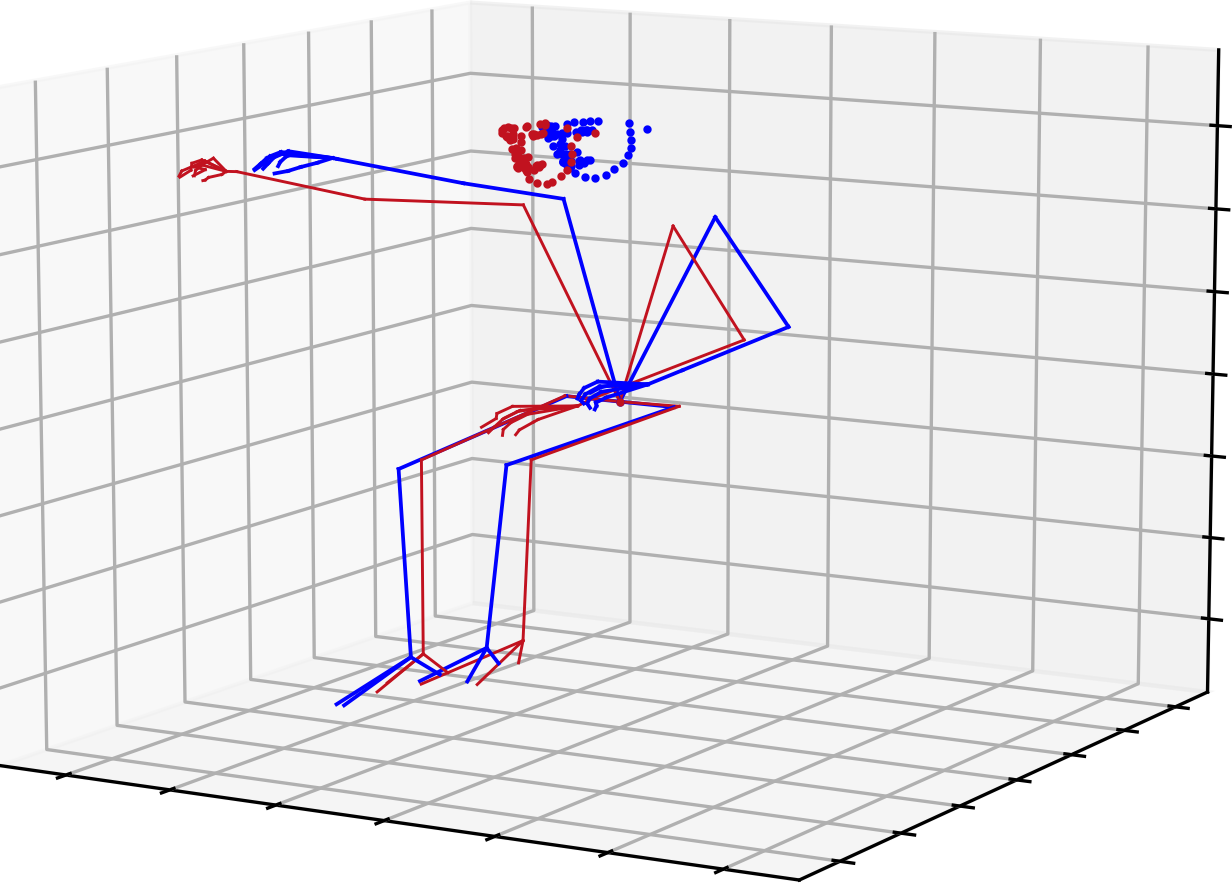} & 
\includegraphics[width=0.18\textwidth,keepaspectratio,] {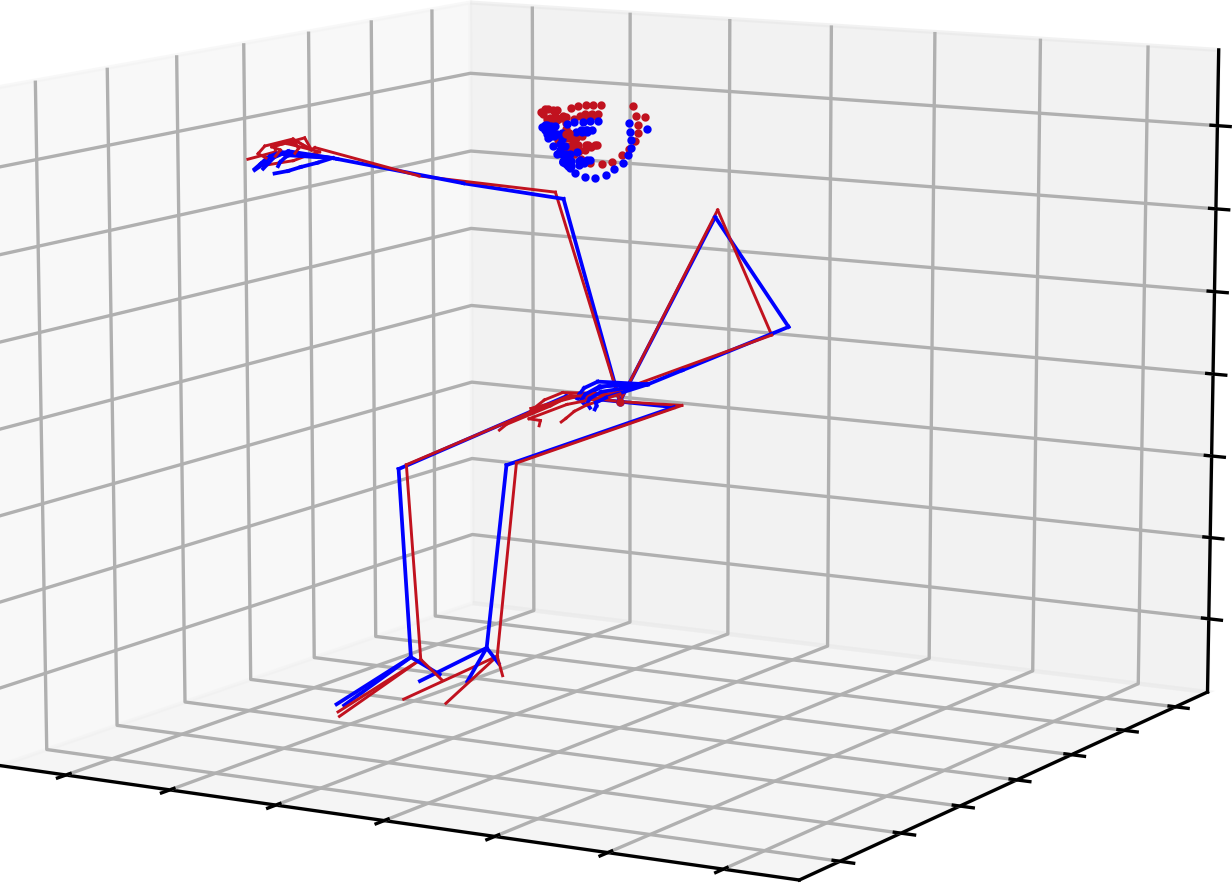}  \\

\includegraphics[width=0.12\textwidth,keepaspectratio,] {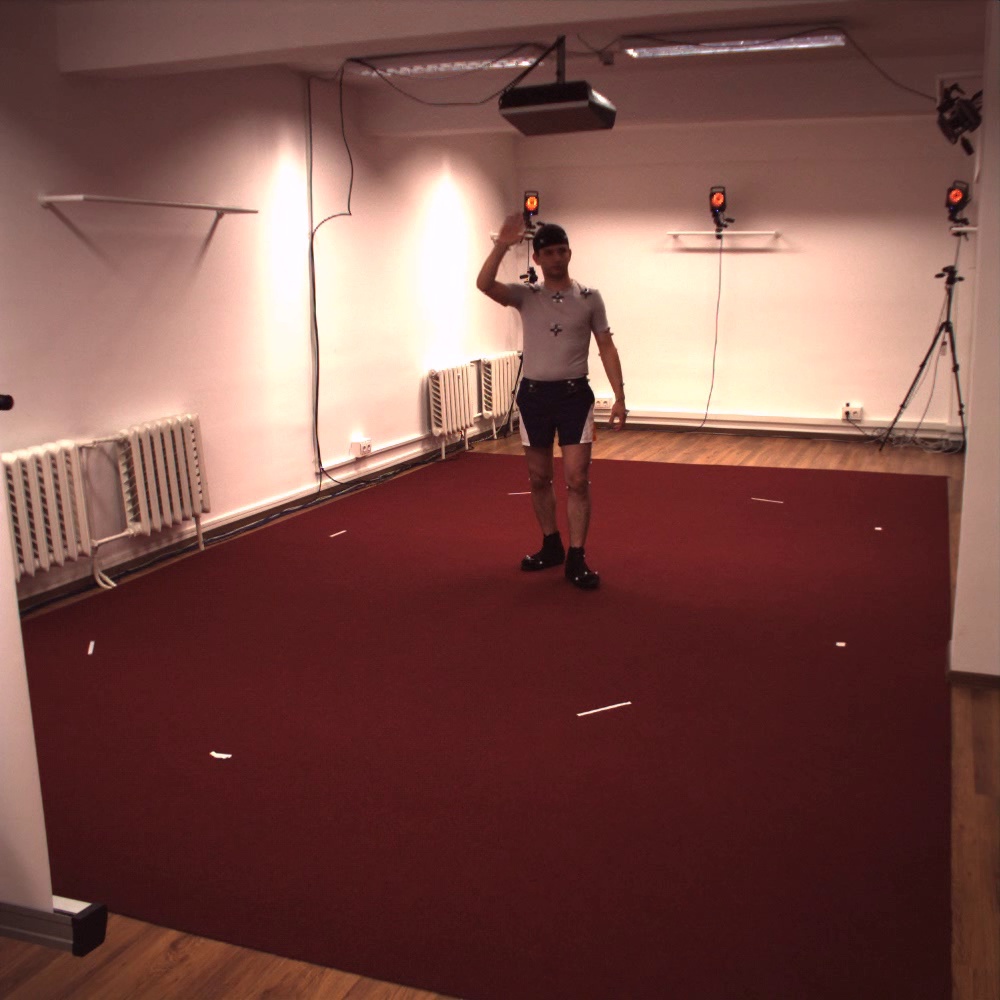} & 
\includegraphics[width=0.18\textwidth,keepaspectratio,] {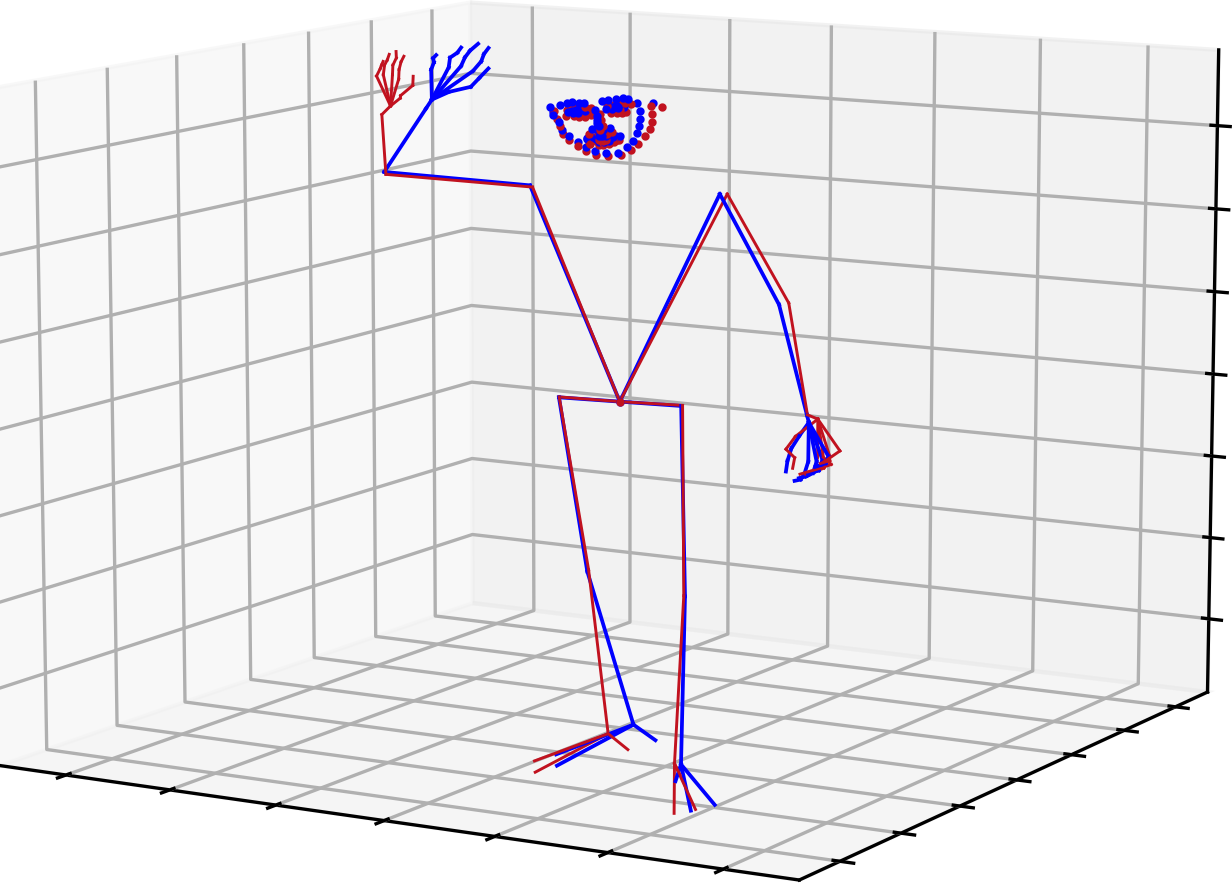} & 
\includegraphics[width=0.18\textwidth,keepaspectratio,] {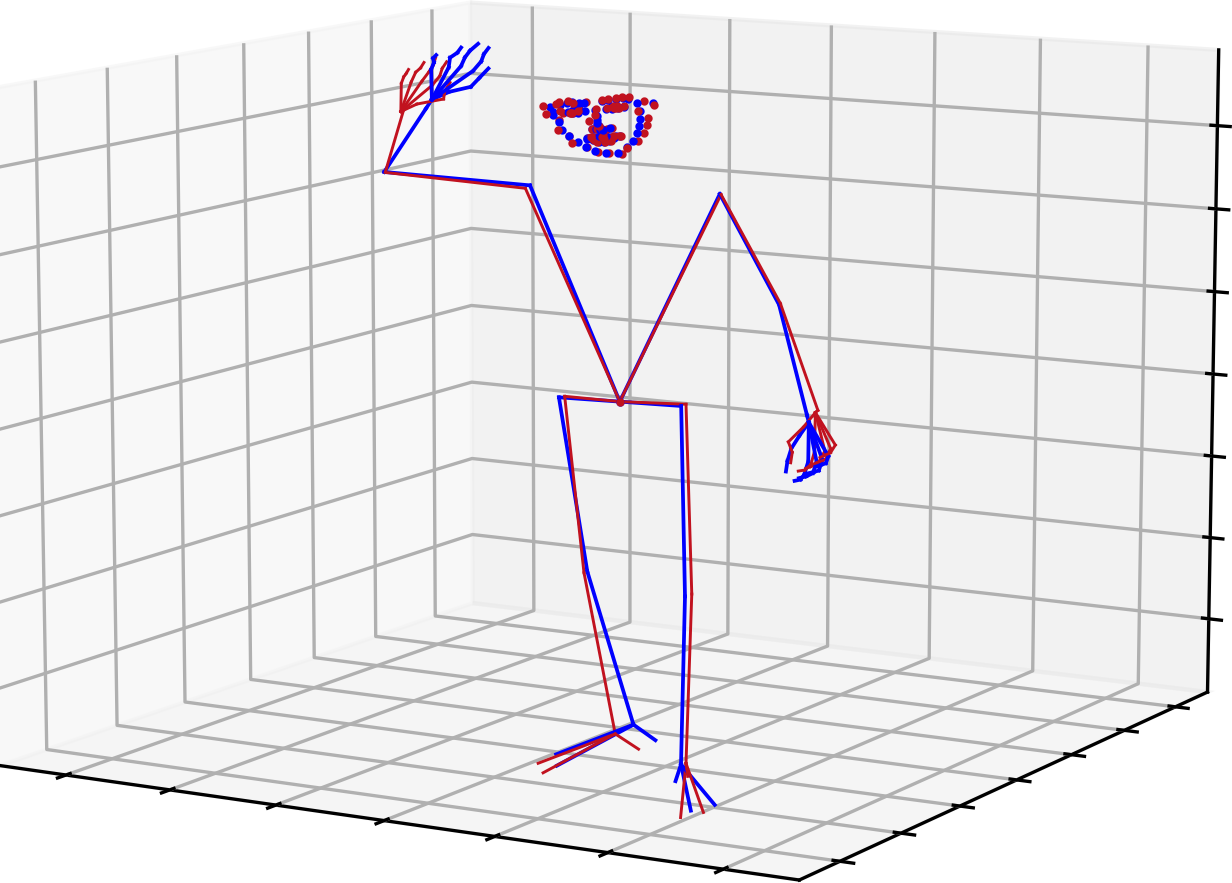}\\

\end{tabular}}
\caption{Qualitative results from the H3WB test set. Blue: ground-truth, Red: best hypothesis. In comparison to D3DP, \ours's is better-aligned to the body joints (\eg, the shoulders), due to the hierarchical structure of the part-based prediction inducing such alignment.
Remark also that \ours's dedicated networks for hands and face lead to considerably better predictions for those body parts.}
\label{fig:visuals_h3wb}
\end{figure}

\begin{figure}[!htb]
 \captionsetup[subfigure]{labelformat=empty}
\centering
\setlength\tabcolsep{1.5pt} 
\resizebox{0.95\textwidth}{!}{
\begin{tabular}{cccc}

\includegraphics[width=0.13\textwidth,keepaspectratio,] {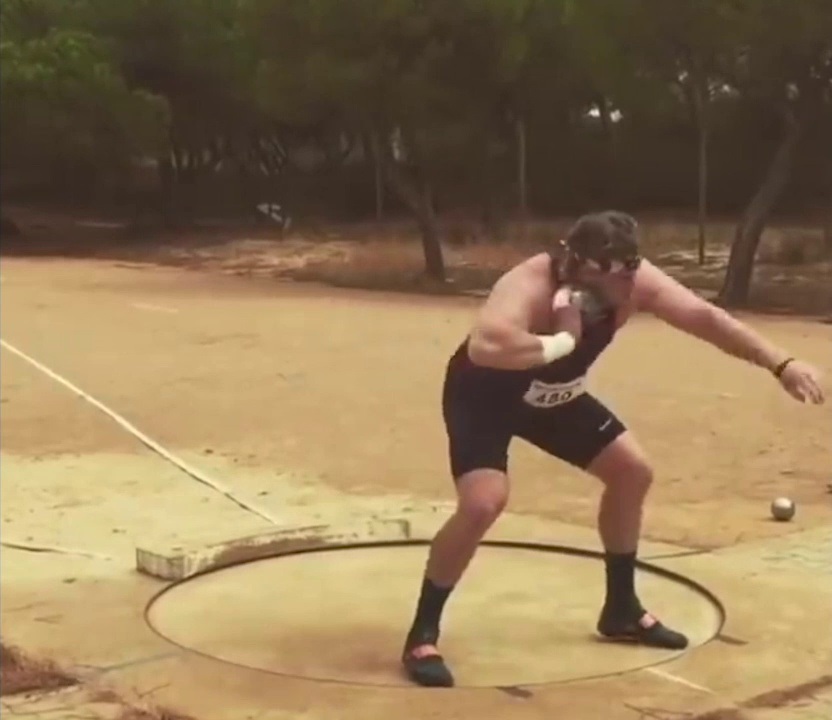} &

\includegraphics[width=0.14\textwidth,keepaspectratio,] {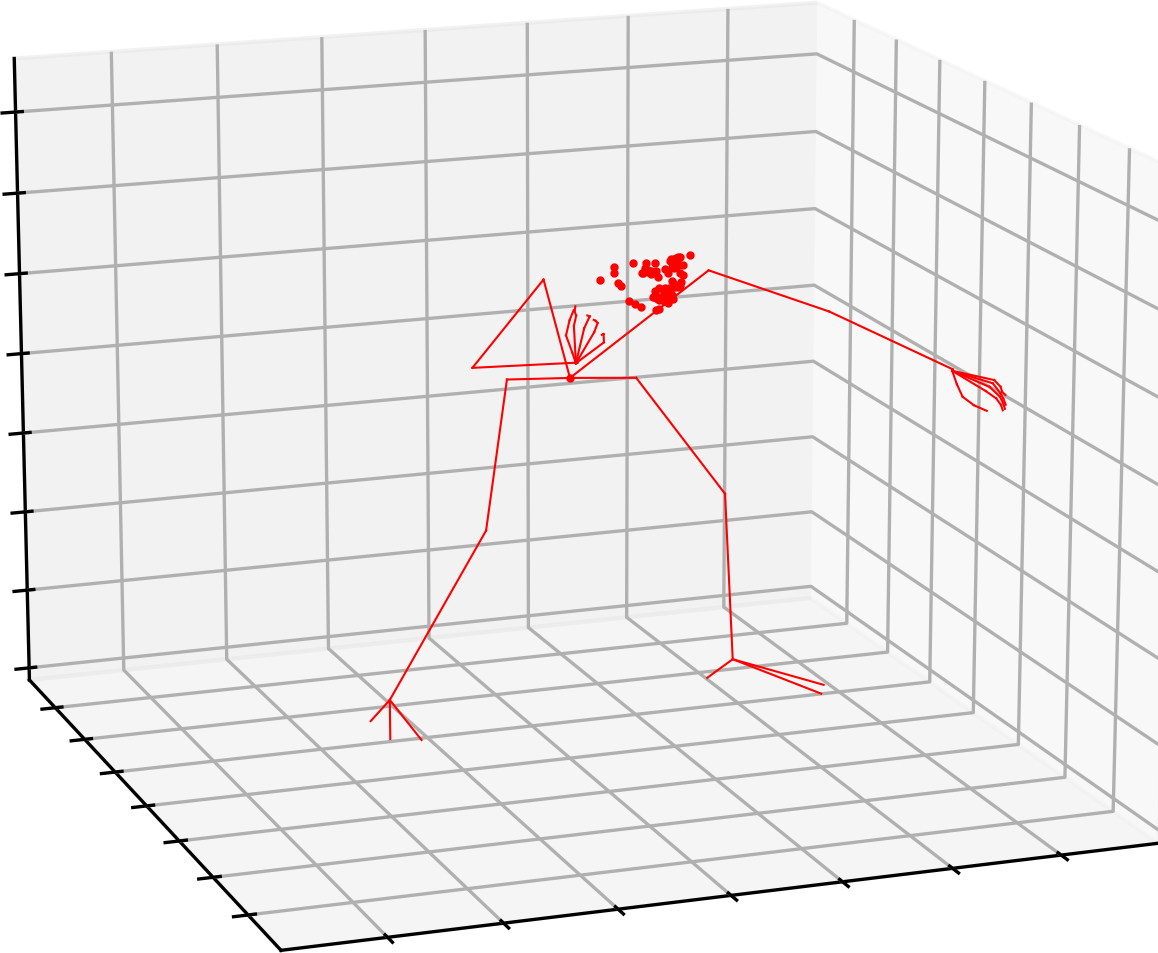} &

\includegraphics[width=0.13\textwidth,keepaspectratio,] {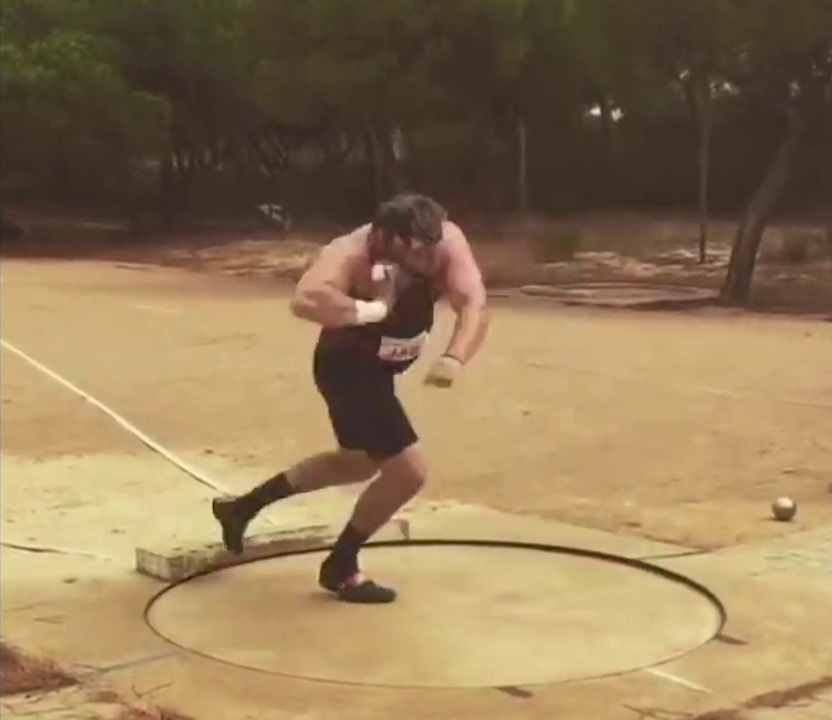} &

\includegraphics[width=0.14\textwidth,keepaspectratio,] {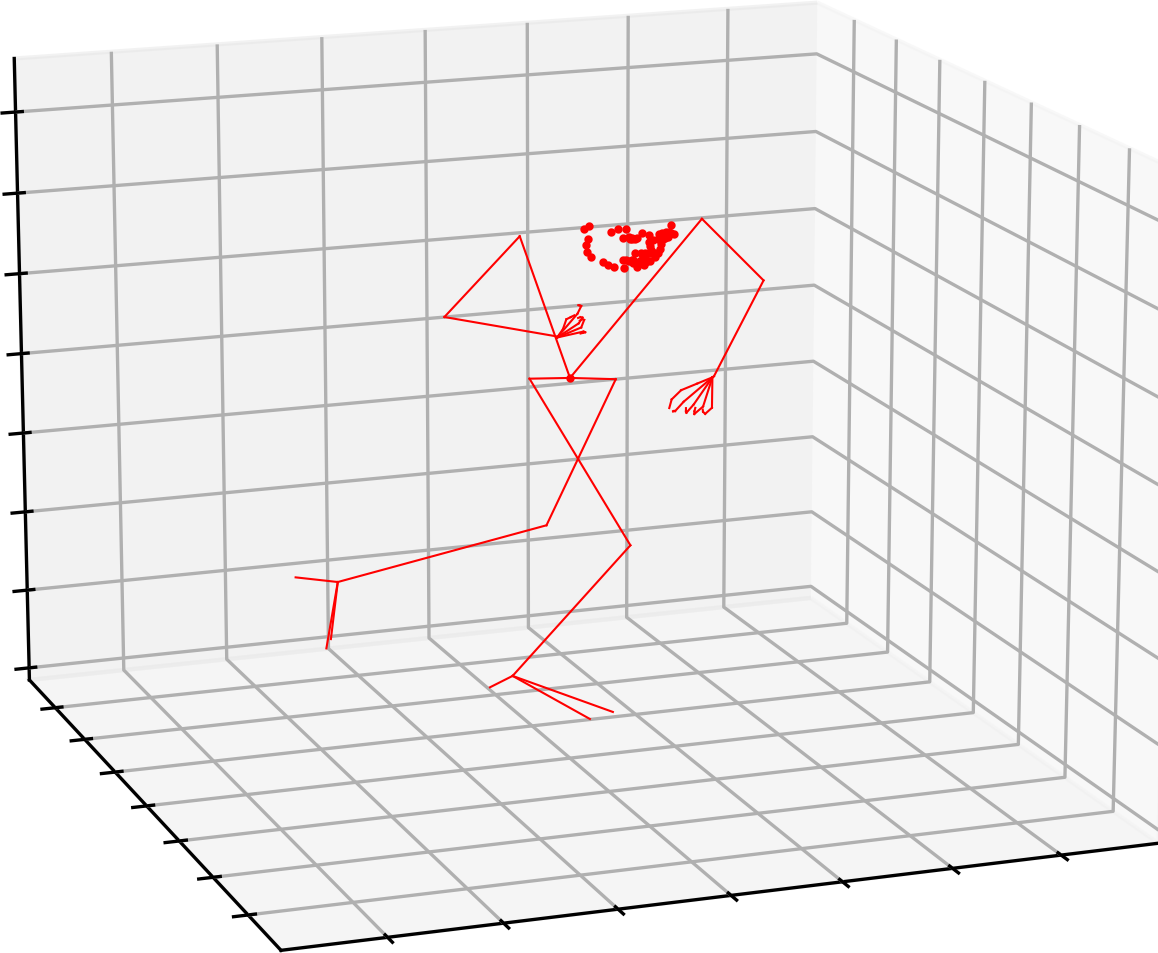} \\

\includegraphics[width=0.18\textwidth,keepaspectratio,] {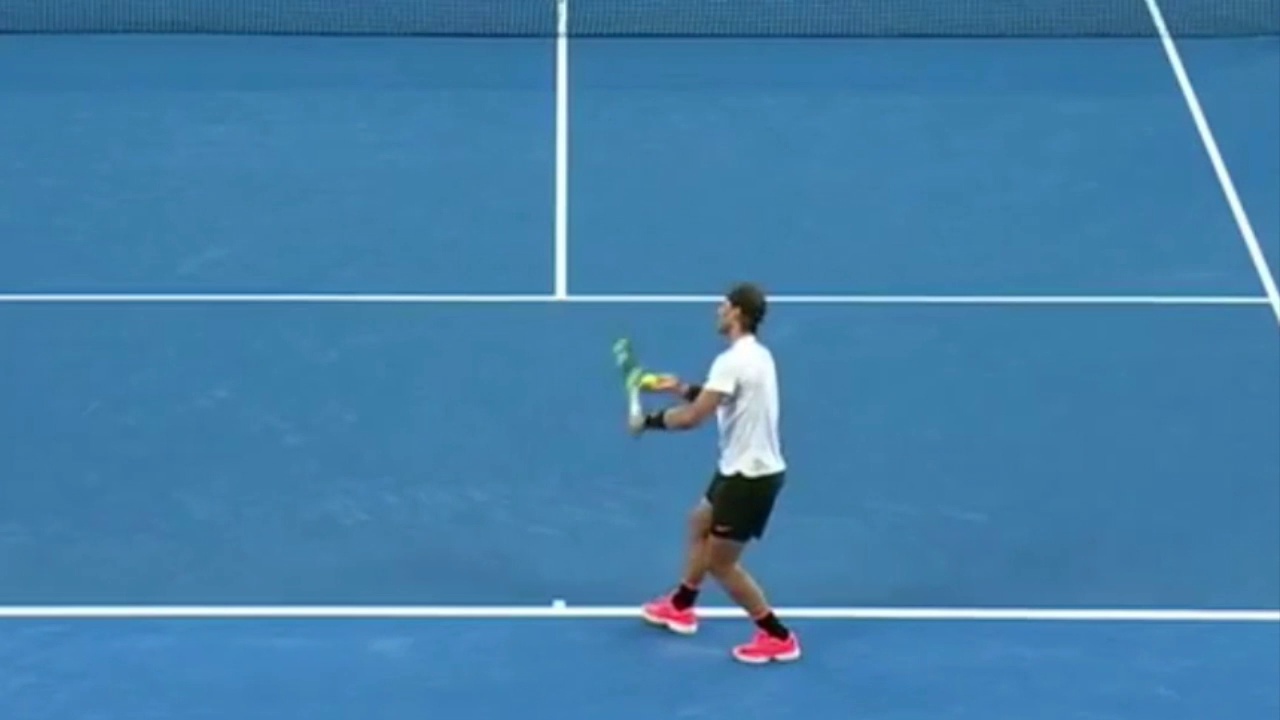} & 

\includegraphics[width=0.14\textwidth,keepaspectratio,] {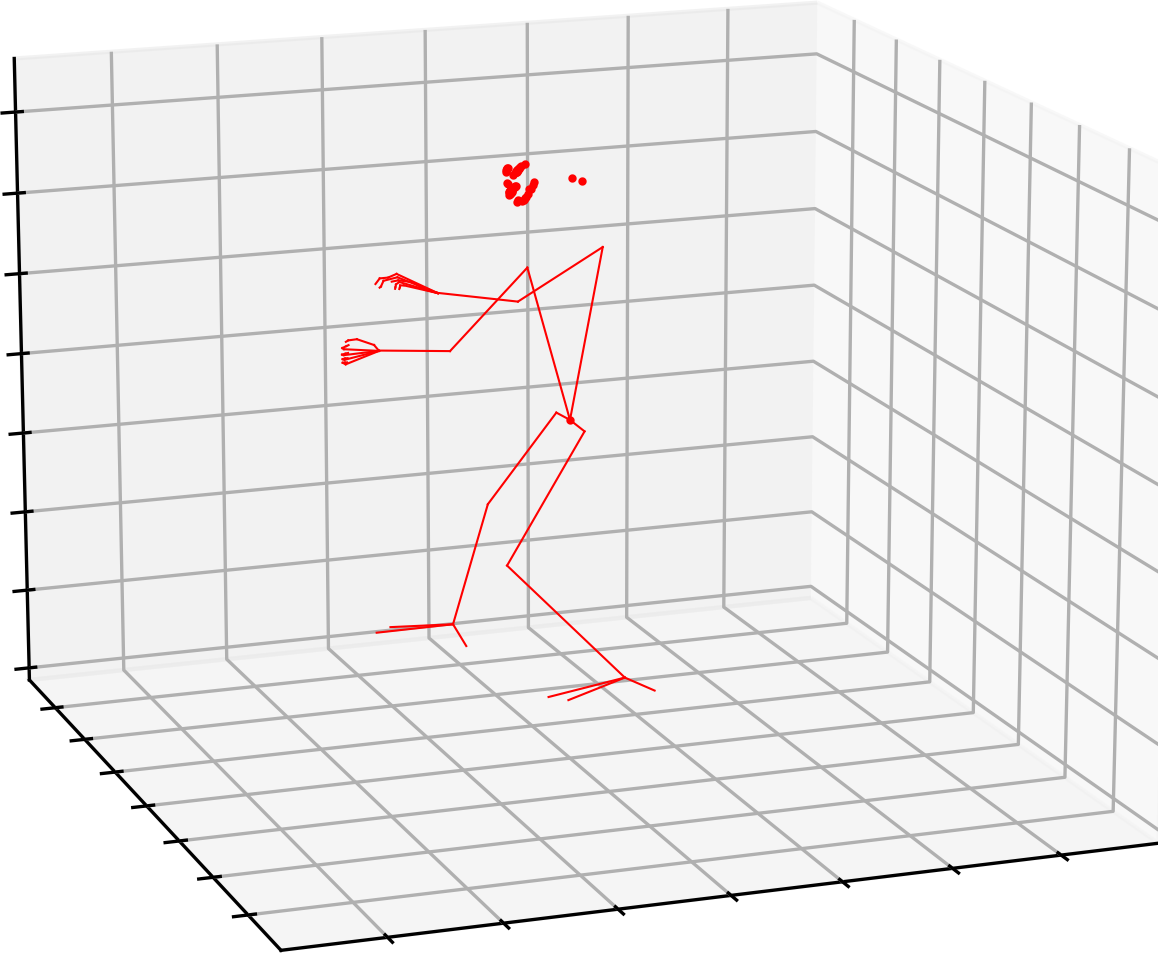} &

\includegraphics[width=0.18\textwidth,keepaspectratio,] {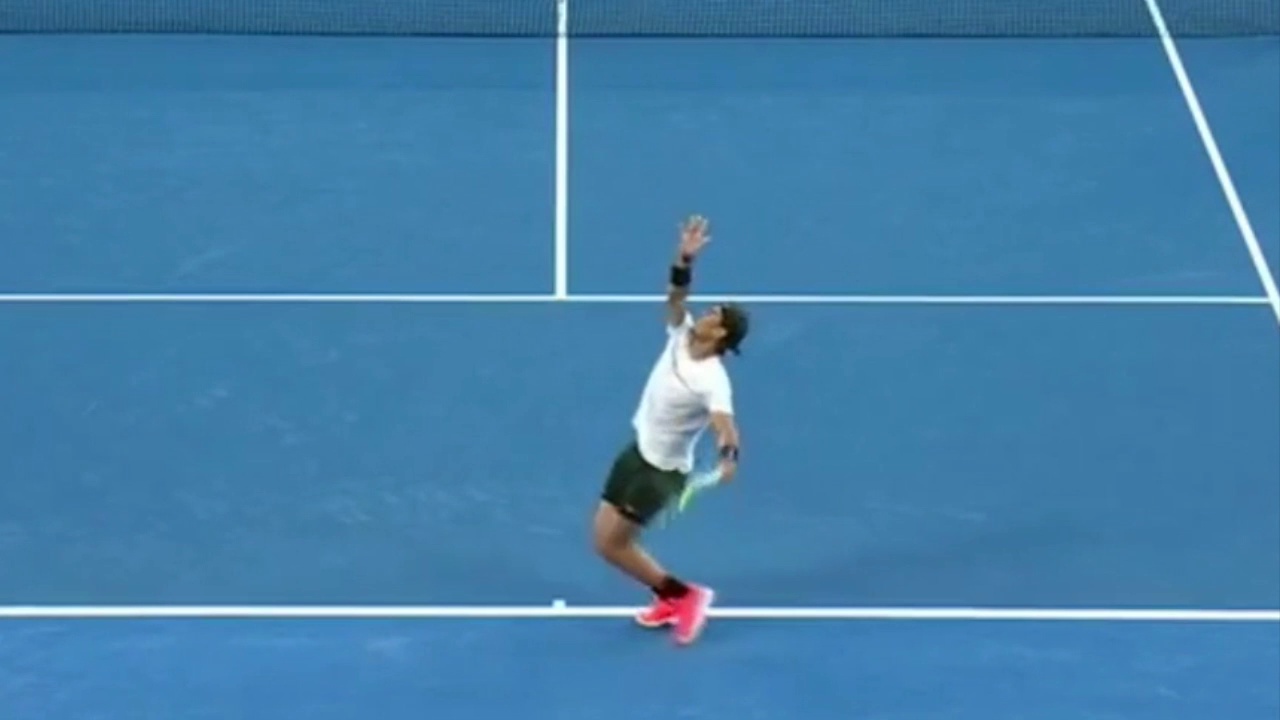} & 

\includegraphics[width=0.14\textwidth,keepaspectratio,] {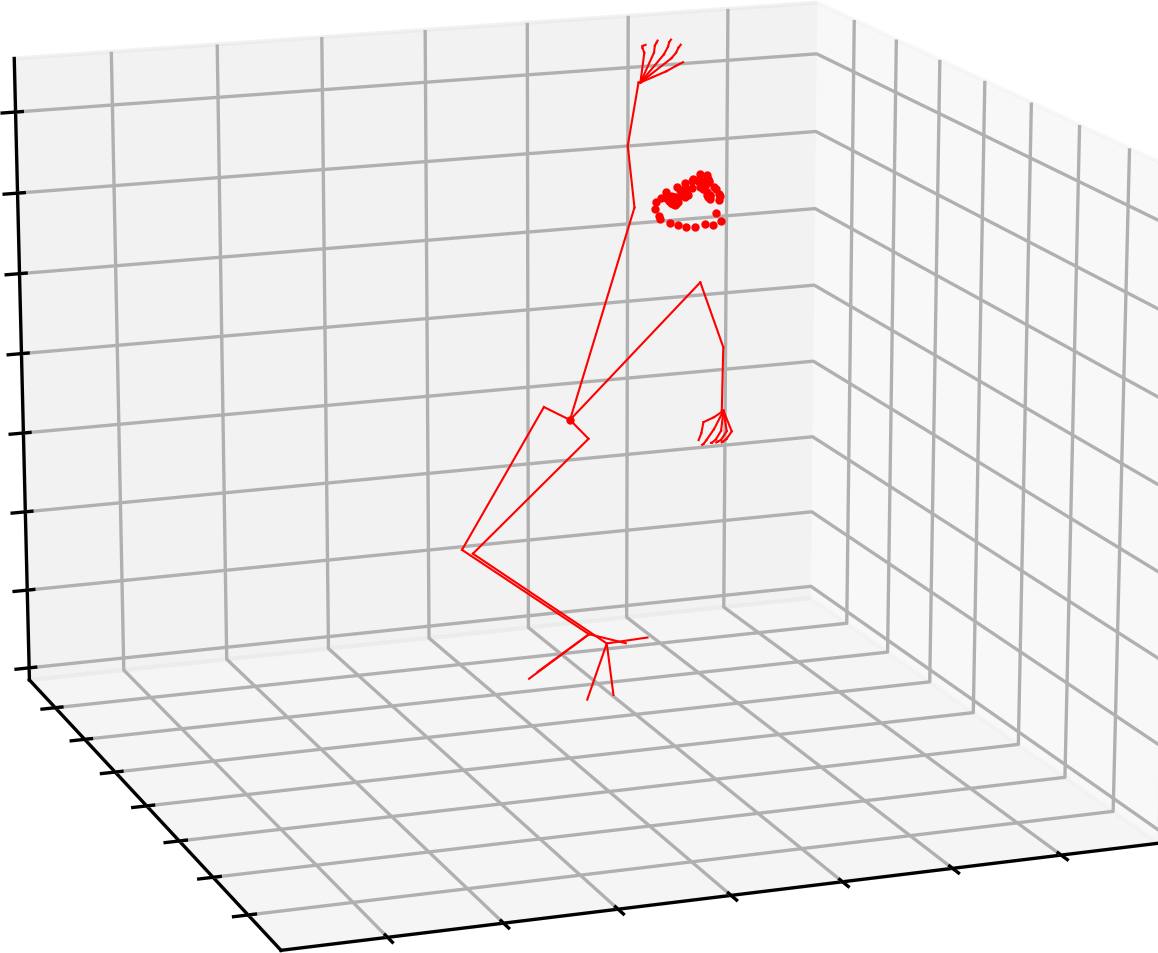} \\

\includegraphics[width=0.18\textwidth,keepaspectratio,] {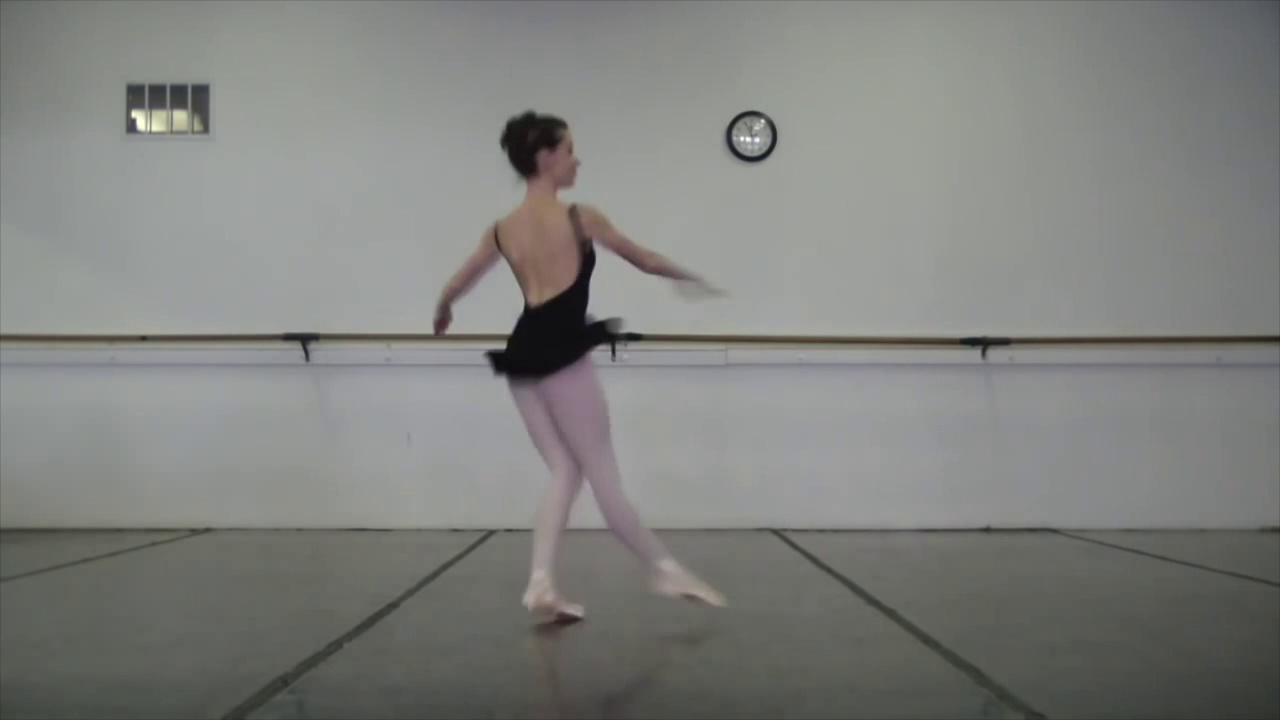} & 

\includegraphics[width=0.14\textwidth,keepaspectratio,] {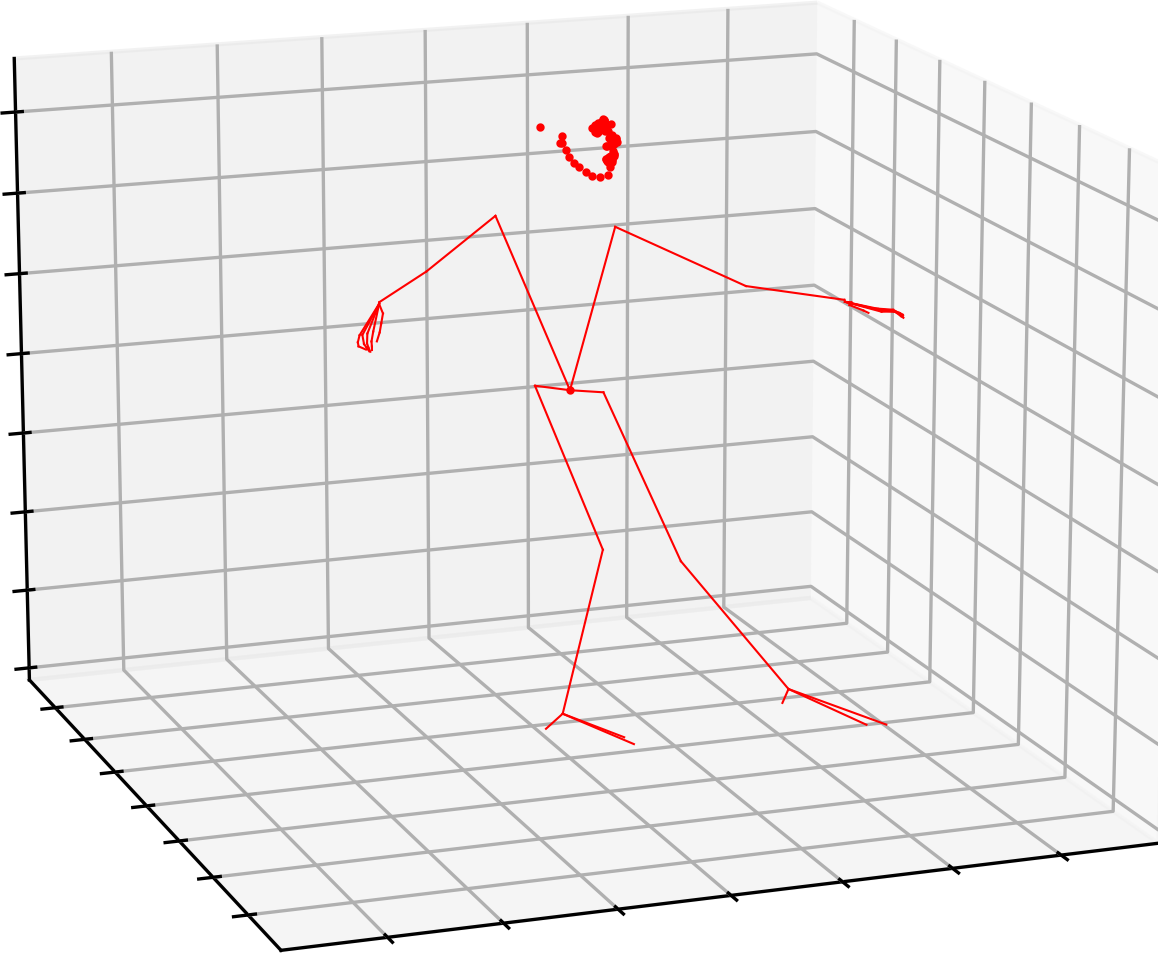} &

\includegraphics[width=0.18\textwidth,keepaspectratio,] {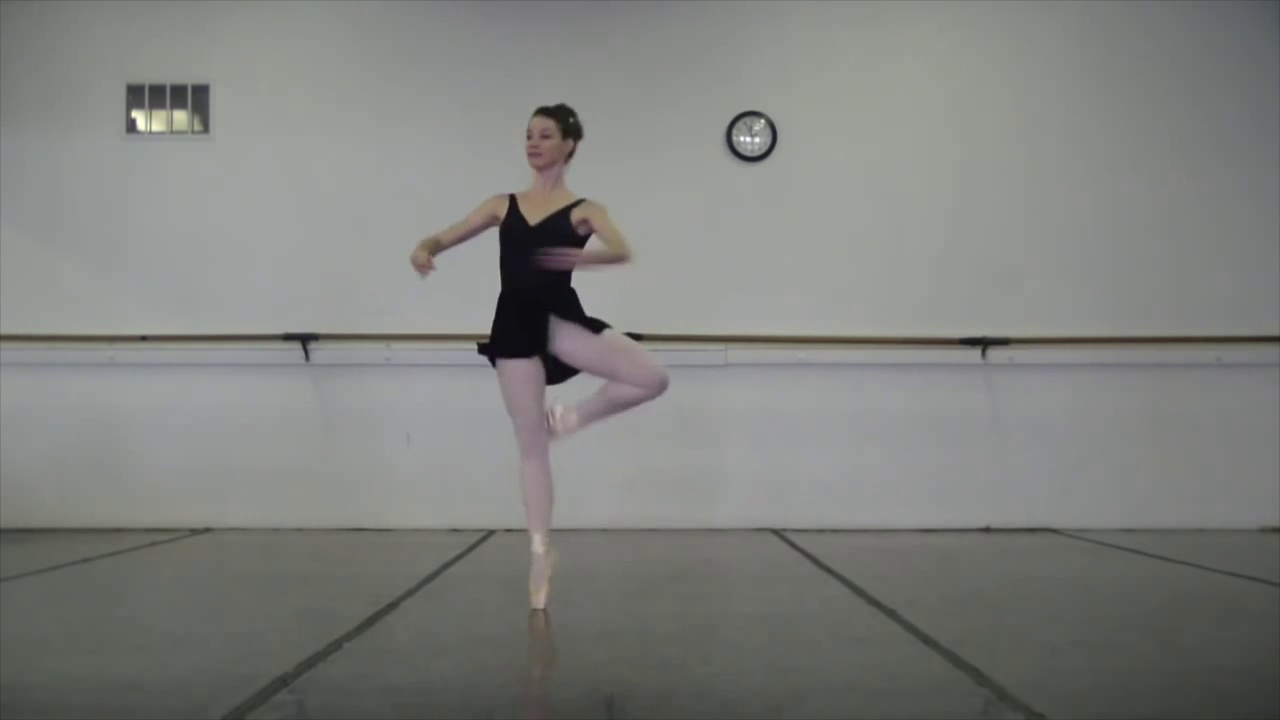} & 

\includegraphics[width=0.14\textwidth,keepaspectratio,] {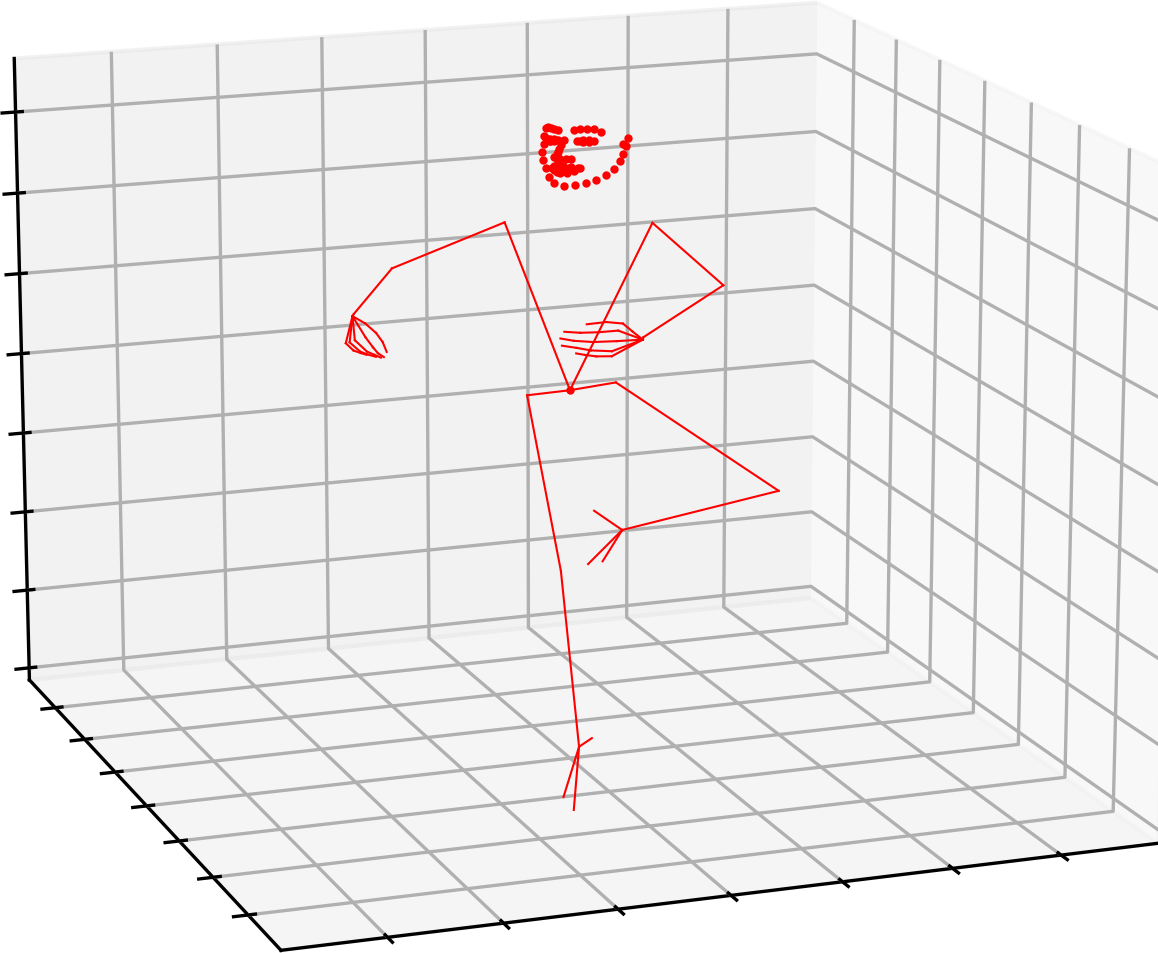}\\

\includegraphics[width=0.18\textwidth,keepaspectratio,] {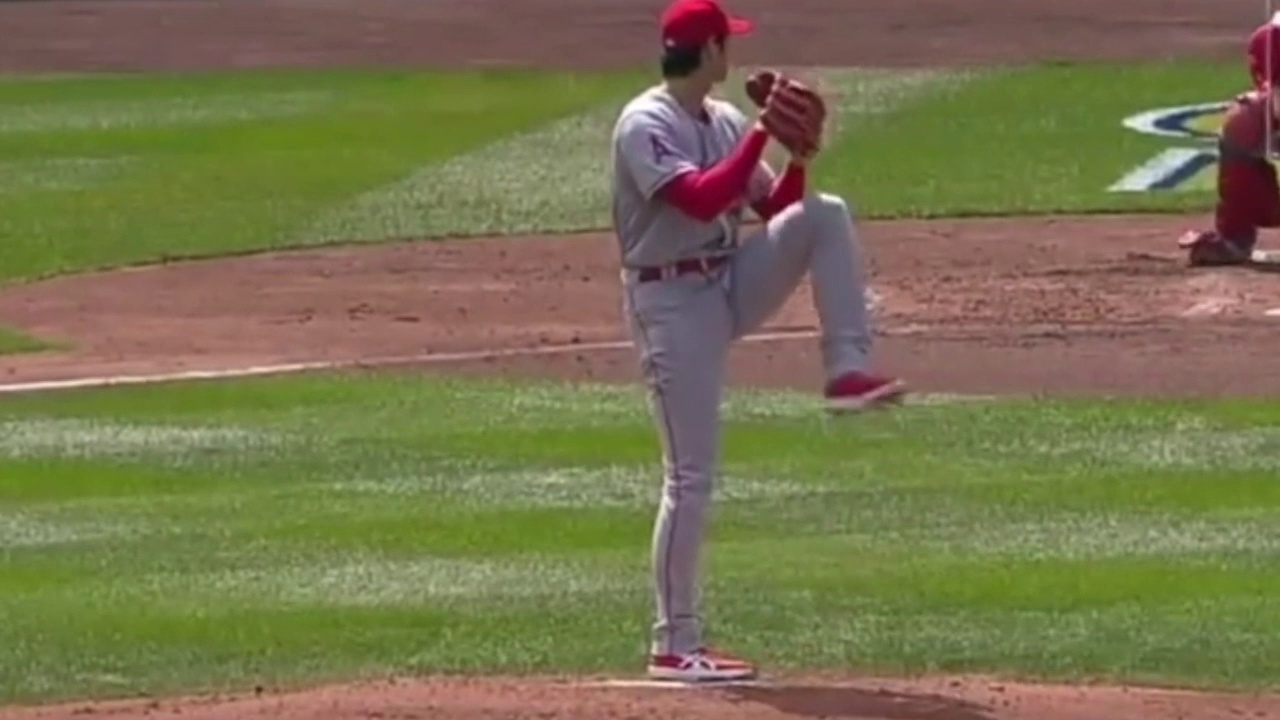} & 

\includegraphics[width=0.14\textwidth,keepaspectratio,] {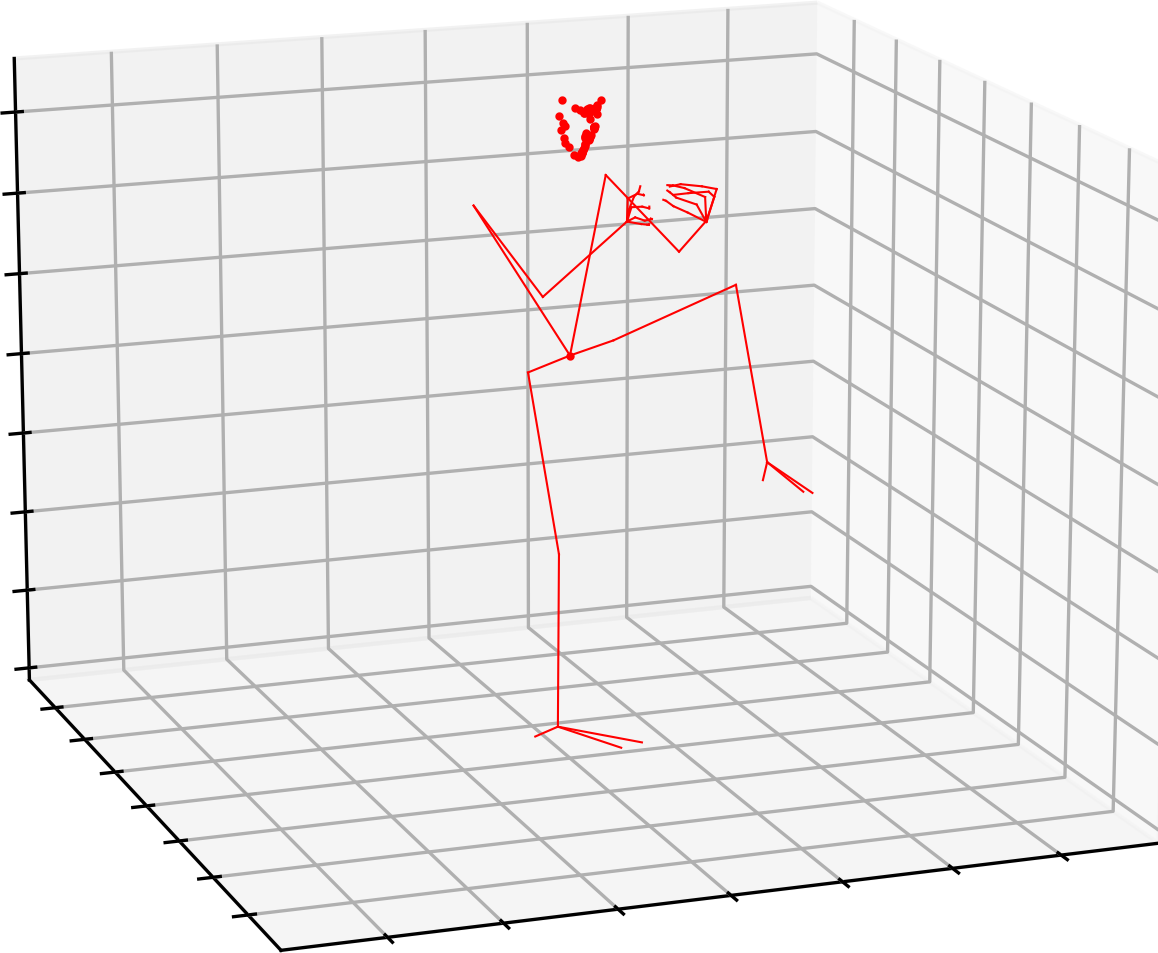} &

\includegraphics[width=0.18\textwidth,keepaspectratio,] {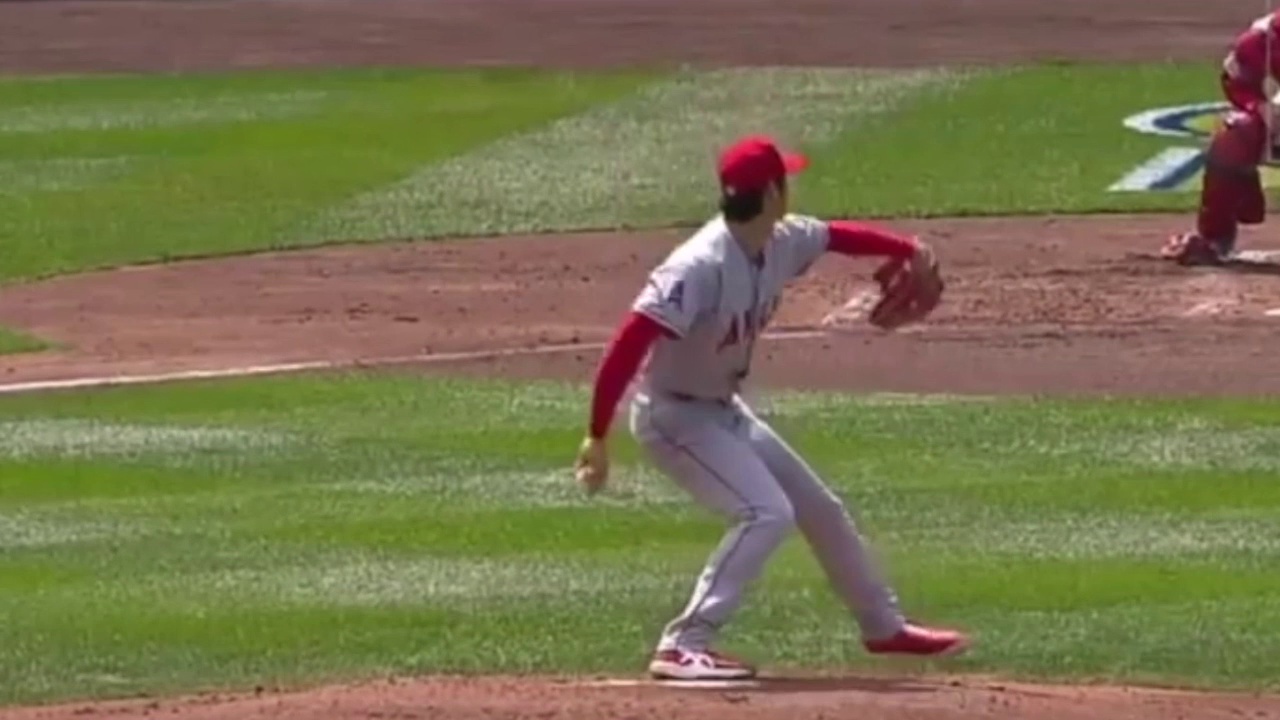} & 

\includegraphics[width=0.14\textwidth,keepaspectratio,] {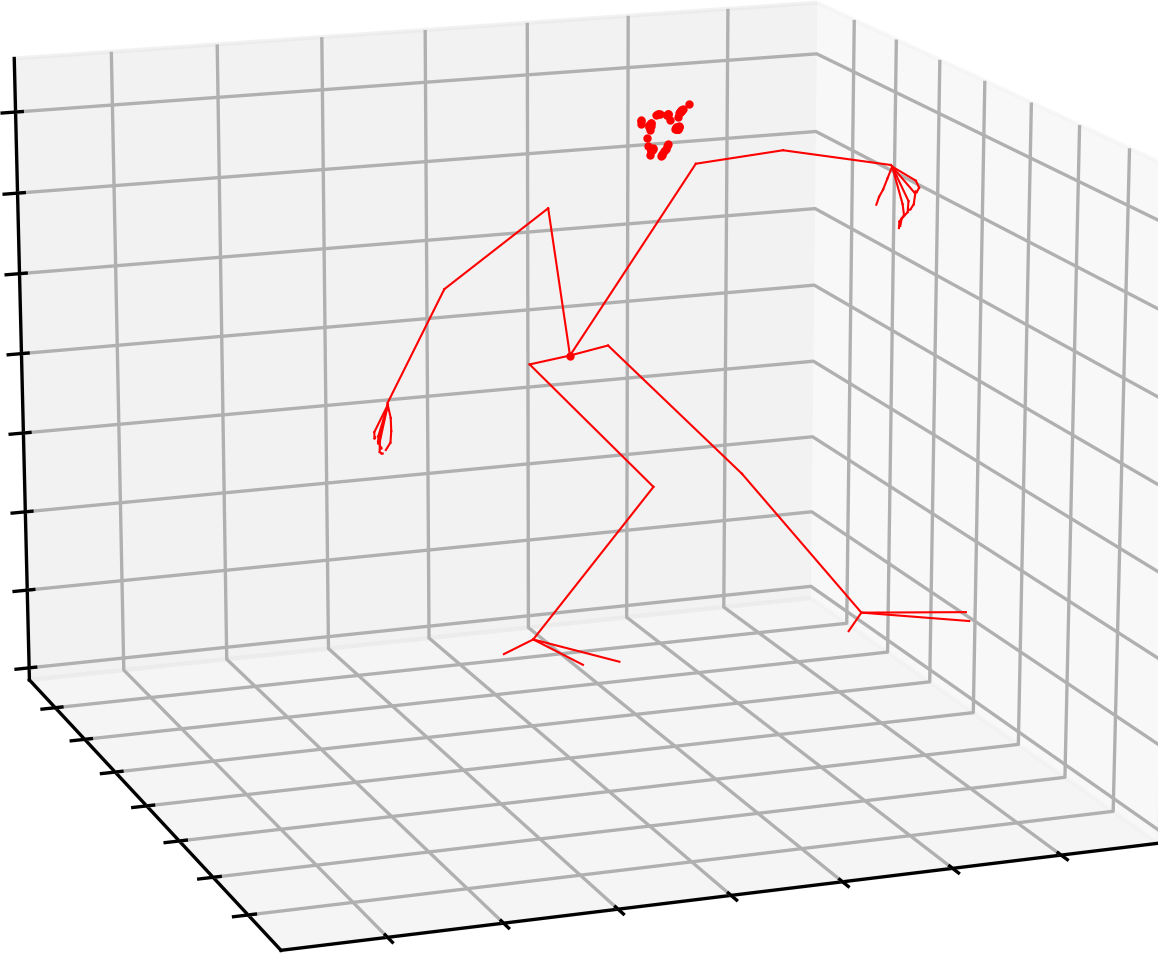}\\

\includegraphics[width=0.18\textwidth,keepaspectratio,] {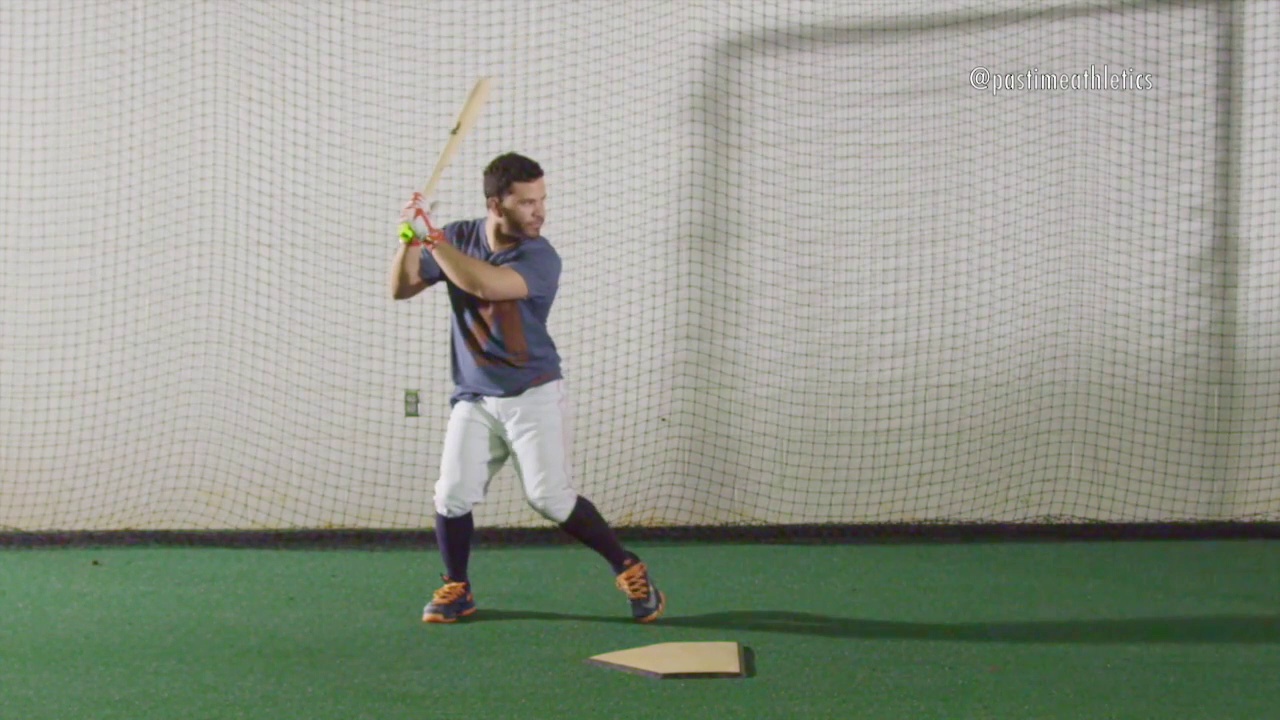} & 

\includegraphics[width=0.14\textwidth,keepaspectratio,] {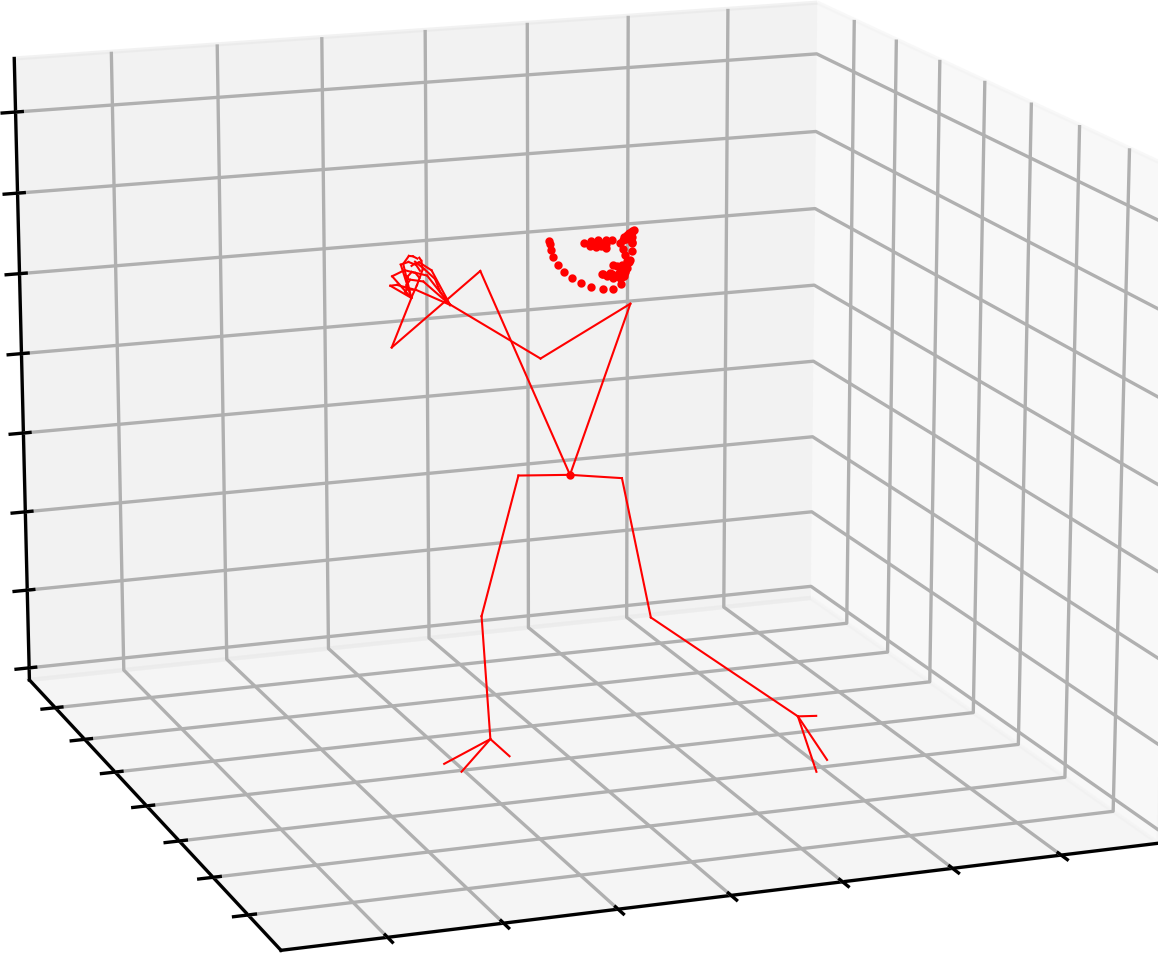} &

\includegraphics[width=0.18\textwidth,keepaspectratio,] {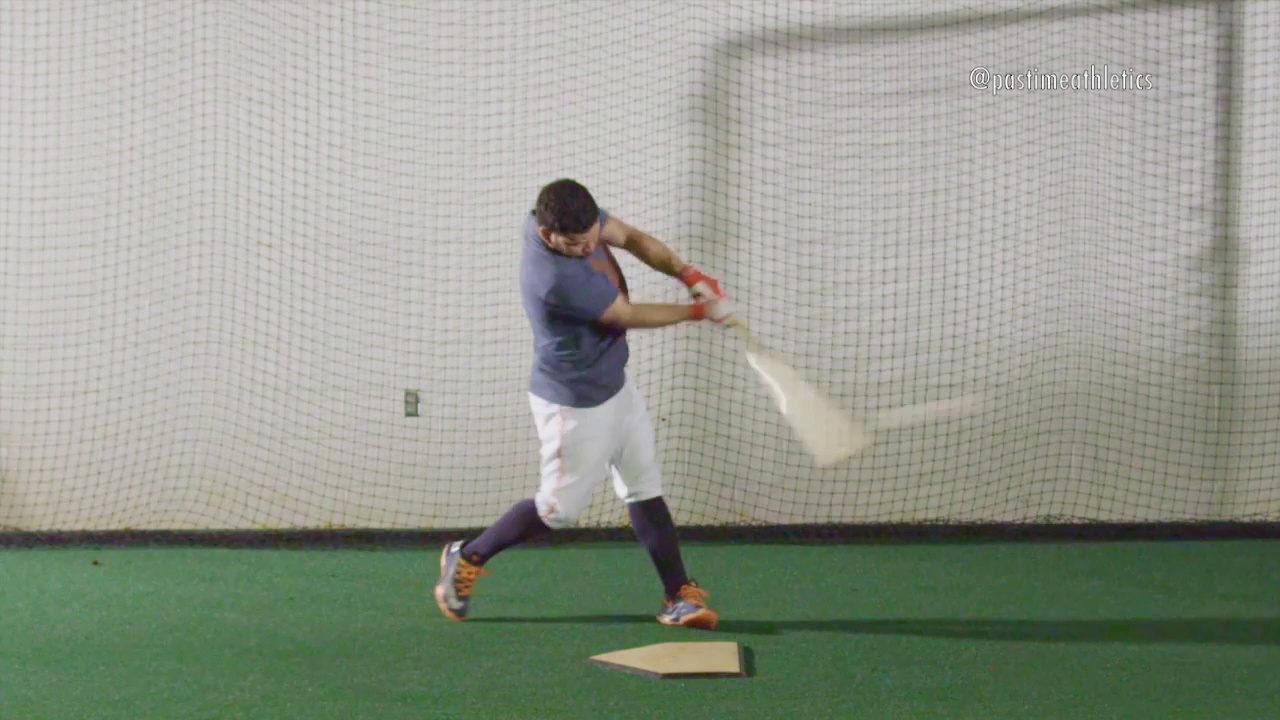} & 

\includegraphics[width=0.14\textwidth,keepaspectratio,] {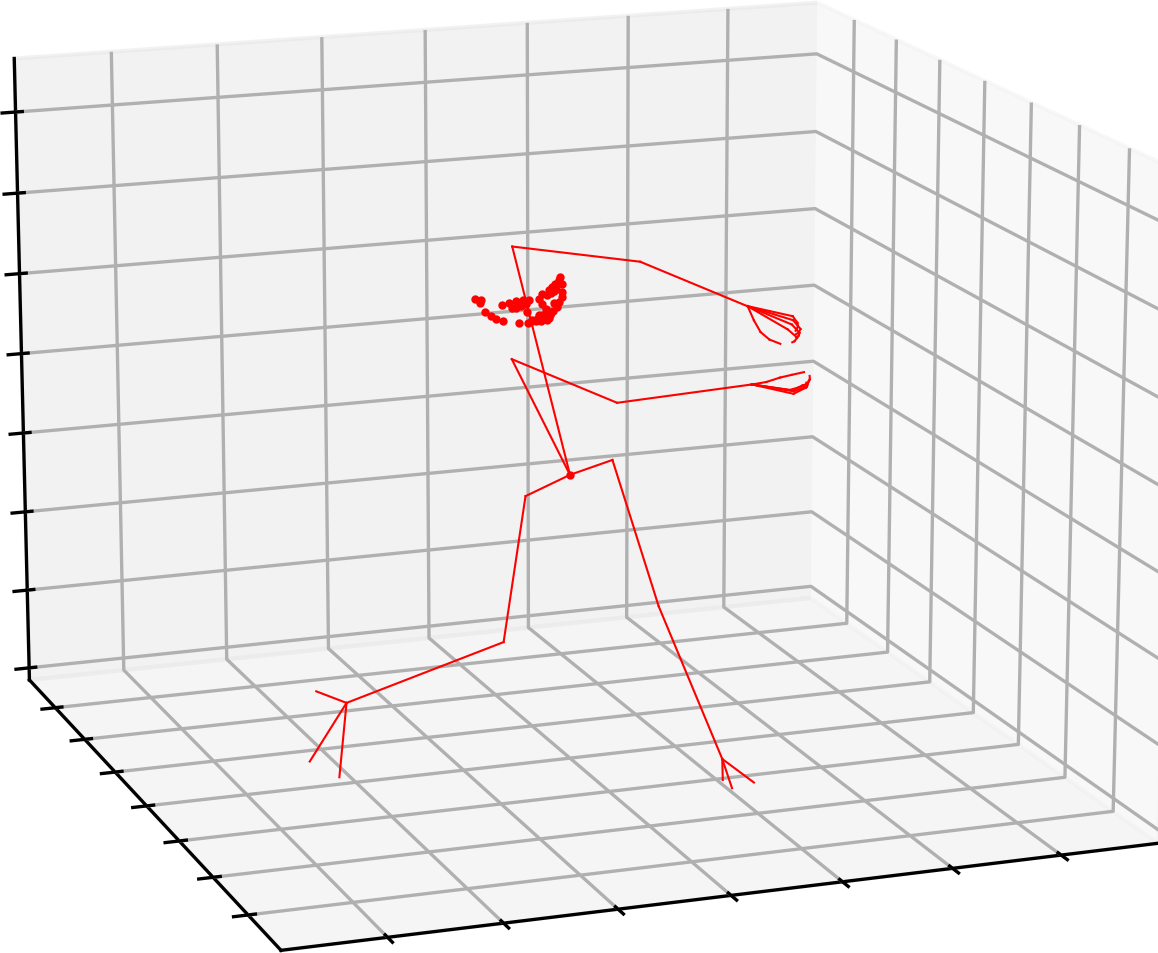} \\

\end{tabular}}
\caption{Qualitative results in the wild. \ours performs well even on challenging scenarios of fast movement, occlusions, and self occlusions. Remark, in particular, the occlusions caused by external objects such as balls (first row), tennis racket (second row), gloves (4th row) and club (last row). Remark also its robustness against motion blur on body parts (ballerina's hands on 3rd row). 2D whole-body keypoints extracted using OpenPifPaf~\cite{pifpaf} and 3D poses predicted by \ours.}
\vspace{-0.5cm}
\label{fig:visuals_wild}
\end{figure}

\paragraph{Qualitative evaluation.} We present several qualitative evaluations in \cref{fig:visuals_h3wb,fig:visuals_wild}. \cref{fig:visuals_h3wb} shows visual results from the test set of H3WB dataset. We observe that our method can successfully predict the body, hand and face under challenging deformations. In particular, it tends to better predict body joint that are key for the body parts, like shoulders and knees, which then leads to better aligned extremities like hands and face.
Furthermore to show the robustness of our method we provide visual results from in-the-wild scenarios in \cref{fig:visuals_wild}. Even under challenging conditions such as in row 2 and 3, our method predicts the corresponding 3D poses of the hardly visible left hand of the tennis player or the blurry right hand of ballerina. Notice also that the cameras of these in the wild scenario are very different from the ones of H3WB (different position, focal length, fov, etc) and yet our prediction are well aligned with the whole-body pose of the subject. We provide more in-the-wild qualitative results in the supplementary material.

\setlength\intextsep{8mm}

\setlength\intextsep{8mm}

\begin{table}
    \centering
    \setlength{\tabcolsep}{5pt}
    \resizebox{0.99\linewidth}{!}{
    \begin{tabular}{l c c c c c c c c}
     \toprule
    Method & Shift & Denoisers & WB & PB &
    Body & Face  &  Hands  \\ 
    \midrule
    D3DP~\cite{d3dp} (baseline) & \xmark & \xmark & 46.9 & 19.4 & 44.6 &  7.9 &  23.6 \\ 
 
    Only part-frame shift& \cmark &  \xmark & 95.5 & 33.6 & 123.1 & 7.7 & 24.6 \\ 
    
    Only part denoisers & \xmark &  \cmark & 89.6 & 21.6 & 41.2 & 10.1 & 29.2 \\ 
    
    \ours & \cmark &  \cmark & \textbf{43.0} & \textbf{16.4} & \textbf{37.1} & \textbf{5.8} & \textbf{21.7} \\
     \bottomrule
    \end{tabular} 
    }
    \caption{Ablation on main components of \ours, all models with the same total number of parameters (MPJPE metric in mm). Both components are crucial to \ours's performance.}
    \label{tab:components}
    \vspace{-1.5cm}
\end{table}

\subsection{Ablations}
\label{sec:ablations}

Unless specified otherwise, we set the \ours's parameters as $N=27$, $H=20$, and $K=10$ in the ablations, and measure the metrics with the P-Best protocol.

\paragraph{Contribution of components.} The impact of each key element introduced in \ours is dissected in Table~\ref{tab:components}, where the baseline is a D3DP~\cite{d3dp} minimally adapted to accept 133 whole-body keypoints.

The two main components of the part-based approach proposed in \ours are the part-frame shift of keypoints, which represents their position in a frame local to their local root joints (\cref{sec:method}), and the part-specific denoisers, which employs three separate denoising models for each part (hands, face, and main body). In the ablation with \textit{only part-frame shift}, we employ a single 512-channel MixSTE denoising backbone, with a weighted loss to compensate the imbalance in the amount of keypoints across body parts. In the ablation with \textit{only part denoisers}, all parts are frame-shifted to a single root joint (keypoint 0), but we reintroduce the three part-specific denoisers of 384, 224, and 256 channels, respectively, for the body, face, and hands. Remark that all models in Table~\ref{tab:components} have \textbf{essentially the same total number of parameters.}

Our ablation shows that both contributions are essential for improving the results over D3DP. This is expected and fairly intuitive as both steps are complementary. Indeed, using only the part-frame shift with a single network introduces a bias that makes, for example, the hands closer to the hip than to the elbow because of this centering. This can create spurious correlation between unrelated keypoints as everything is processed by a single network. These spurious correlations are removed when separating the networks since the centered keypoints are now processed independently. Similarly, using separated body-part networks without the part-frame shifts forces the hands and the face networks to predict the absolute pose of these parts without having access to the body pose and in particular the pose of their part root joints (wrists and neck), significantly increasing the difficulty of the task.

\begin{table}
\centering
\setlength{\tabcolsep}{5pt}
\begin{tabular}{lccccc}
\toprule
Model & WB & PB & Body& Face & Hands \\ 
\midrule 
\rowem \ours (with default choices) &  43.0 & 16.4  & 37.1 & 5.8 & 21.7 \\
Part Loss $\rightarrow$ WB Loss  & 47.7 & 21.5 & 50.4 & 7.2 & 28.1 \\
Balanced $\rightarrow$ Uniform Denoisers & 49.9 & 17.8 & 46.0 & 5.2  & 22.1 \\
$N=27 \rightarrow N=81$ & 45.1 & 16.3 & 38.8 & 5.6 & 20.7 \\
MPJPE $\rightarrow$ MSE & 44.1 & 18.1 & 42.1 & 6.5 & 23.2 \\

\bottomrule 
\end{tabular}
\caption{Ablation experiments on design choices, comparing the effect of substituting each choice from the default base model $\rightarrow$ an alternative design possibility. 
}
\label{table:abl_table}
\vspace{-1cm}
\end{table}

\paragraph{Design choices.}
In addition to the main components, Table~\ref{table:abl_table} showcases results related to finer design choices, each row illustrating the impact of changing one design axis by its baseline alternative. All choices of \ours lead to better results, except for the frame window length ($N$), in which there seems to be a compromise between balanced performance (with a shorter window of $N=27$, which is our choice), and emphasizing face and hands performance ($N=81$).

The part-based loss of \cref{eq:loss_part} brings the most dramatic gains, in contrast to the whole-body loss of \cref{eq:loss_wb}, demonstrating that separating the optimization targets for each part results in substantial improvement. Balancing the size of the denoiser models (384, 256 and 224 channels for body, hands, and face respectively) also considerably improved the results in comparison to using an uniform-sized (288 channels) denoiser.
Other design choices brought smaller, but still visible improvements. \textit{N=27} $\rightarrow$ \textit{N=81} experiment shows that extending the temporal context adversely affects performance. We attribute this decline to the uneven sampling characteristics of the H3WB dataset. 
The ablation study comparing \textit{MPJPE} to \textit{MSE} shows that despite the common use of MSE loss in diffusion-based methods for 3D whole-body pose estimation, optimizing with MPJPE yields superior results. This outcome aligns with expectations, given that model performance is evaluated using the MPJPE metric.

\section{Conclusion}
\label{sec:conclusion}
We introduced \ours, a denoising diffusion model that predicts entire temporal sequences of 3D whole-body skeleton from their 2D counterparts. 

While diffusion models are SOTA in 3D pose estimation, making them part-based is well-motivated, as the body, hands, and face present widely different scales and motion ranges. However, designing the best combination of diffusion and part-based is far from trivial, as our ablation experiments showcase.

\ours employs a hierarchical part-based model, where body parts that have different scales and motion are predicted by part-specific networks, conditioned on their parent body parts. We were the first to exploit H3WB~\cite{h3wb} temporal information, obtaining outstanding improvements over the previous frame-only state-of-the art (41.4mm MJPJE against 88mm).

\paragraph{Limitations.} Performing all  quantitative experiments on a single dataset is a necessary limitation, as H3WB is the only publicly annotated dataset for whole-body with sequential frames, allowing us to reconstruct the temporal data. We worked to mitigate that limitation by complementing out method with qualitative results in-the-wild videos, showing that our methods generalizes well to unseen context. We expect that future developments of this exciting research are may bring a growing array of data.

\clearpage
%
%
\bibliographystyle{splncs04}
\bibliography{main}
\end{document}


\title{PAFUSE: Part-based Diffusion for \\ 3D Whole-Body Pose Estimation \\ (\textit{Supplementary Material})}

\author{Nermin Samet\inst{1}\orcidlink{0000-0001-9247-2504} \and
C\'edric Rommel\inst{1}\orcidlink{0000-0002-9416-0288} \and
David Picard\inst{2}\orcidlink{0000-0002-6296-4222} \and
Eduardo Valle\inst{1}\orcidlink{0000-0001-5396-9868}}

\authorrunning{N.~Samet et al.}

\institute{Valeo.ai, Paris, France \and
LIGM, Ecole des Ponts, Univ Gustave Eiffel, CNRS, Marne-la-Vall\'ee, France}

\maketitle

\section{Introduction}

We supplement the main article with details on how we adaptated the spatio-temporal models (VideoPose~\cite{videopose}, PoseFormer~\cite{poseformer}, MixSTE~\cite{mixste} and D3DP~\cite{d3dp}) to work with our part-based approach. 

We also provide full videos corresponding to the in-the-wild results presented in the Figure 6, in addition to 4 extra in-the-wild scenarios.\footnote{The videos and this supplemental document were compressed together in the same zip file.} Our method performs well on challenging scenarios of fast movement, occlusions, and self occlusions. Please remark that some failure cases are not due to PAFUSE, but to the underlying 2D whole-body keypoint extractor (OpenPifPaf~\cite{pifpaf}).

We emphasize that our code will be publicly released upon the publication of the paper.

\paragraph{Technical details on part-based adaption of spatio-temporal models.}
In order to support VideoPose~\cite{videopose}, PoseFormer~\cite{poseformer}, MixSTE~\cite{mixste} and D3DP~\cite{d3dp} for part-based 3D whole-body pose estimation, we adapt each model as follows:
\begin{enumerate}
    \item We create 3 pose estimation networks for body, face and hands, and adapt each network to accept the number of keypoints in their corresponding parts.
    \item Next, we train each network with the part-shifted keypoints on the corresponding body parts.
    \item In order to preserve the original number of parameter for the models, we adjust the number of the input channels in the first  layer of each network for each body model. For instance, in VideoPose~\cite{videopose}, the number of channels of the first layer is $1024$, yielding $9.6M$ trainable parameters for whole-body pose estimation (directly with 133 keypoints). In the part-based model, we set number of channels to $1024$, $256$, $256$ for the body, face, and hand networks, respectively, yielding $9.9M$ trainable parameters. We present detailed comparisons for each model in Table~\ref{tab:details}.
\end{enumerate}

\begin{table}
    \centering
    \setlength{\tabcolsep}{5pt}
    \begin{tabular}{l c c c c}
    \toprule
    
    Model & \shortstack{WB Embed. \\ Dim. Size}  & \shortstack{PB Embed. Dim. Size \\ \textit{body} / \textit{face} / \textit{hand}}  & \shortstack{WB Num. \\ Params} & \shortstack{PB Num. \\ Params} \\
    \midrule
    VideoPose~\cite{videopose} & $1024$ & $1024\text{ }/\text{ }256\text{ }/ 256$ & $9.6M$ & $9.9M$ \\
    PoseFormer~\cite{poseformer} &  $32$ & $128\text{ }/\text{ }32\text{ }/ 48$ & $59.1M$ & $58.5M$\\
    MixSTE~\cite{mixste} & $512$ & $384\text{ }/\text{ }224\text{ }/\text{ } 256$ & $33.7M$ & $33.9M$ \\
    D3DP~\cite{d3dp} & $512$ & $384\text{ }/\text{ }224\text{ }/\text{ }256$ & $34.8M$ & $34.9M$ \\
 
     \bottomrule
    \end{tabular}
    
    \caption{Detailed comparison of whole-body and part-based pose estimation models, showing input layer embedding dimension and total number of parameters. WB: Whole-Body. PB: Part-Based.}
    \label{tab:details}
    
\end{table}

\newpage

\bibliographystyle{splncs04}
\bibliography{main}